\documentclass[twoside,11pt]{article}

\usepackage{jmlr2e}

\newcommand{\bx}{\mathbf{x}}

\newcommand{\bz}{\mathbf{z}}

\newcommand{\ccV}{\mathcal{V}}
\newcommand{\ccZ}{\mathcal{Z}}

\newcommand{\ccC}{\mathcal{C}}

\newcommand{\kk}{$(k\!-\!1)$}
\newcommand{\kks}{$(k\!-\!1)$ }

\newcommand{\bu}{\mathbf{u}}

\newcommand{\bw}{\mathbf{w}}

\newcommand{\bth}{\boldsymbol{\theta}}

\newcommand{\eqq}{$\,=\,$}
\newcommand{\no}{\noindent}

\newcommand{\ben}{\begin{eqnarray*}}
\newcommand{\een}{\end{eqnarray*}}
\newcommand{\be}{\begin{eqnarray}}
\newcommand{\ee}{\end{eqnarray}}

\newcommand{\blll}{\hspace*{-3.5mm}{\tiny $\bullet$} }
\newcommand{\cca}{ {\cal C}}

\newcommand{\ts}{\text{s}}

\jmlrheading{1}{2020}{20-48}{2/20}{}{Bahman Moraffah and Antonia Papandreou-Suppappola}

\ShortHeadings{BNP Modeling for Multiple Object Tracking}{Moraffah and Papandreou-Suppappola}
\firstpageno{1}

\begin{document}

\title{Bayesian Nonparametric Modeling for Predicting Dynamic Dependencies in Multiple Object Trackings}

\author{\name Bahman Moraffah \email bahman.moraffah@asu.edu \\
       \addr Department of Electrical Engineering\\
       Arizona State University\\
       Tempe, AZ 85281, USA
       \AND
       \name Antonia Papandreou-Suppappola \email papandreou@asu.edu \\
       \addr Department of Electrical Engineering\\
       Arizona State University\\
      Tempe, AZ 85281, USA}

\editor{}

\maketitle

\begin{abstract}
Some challenging problems in tracking multiple objects 
include the time dependent cardinality, unordered measurements
and object  parameter labeling. In this paper, we employ 
Bayesian  Bayesian nonparametric methods
to address these challenges. In particular, we propose  
modeling the multiple object parameter state prior using the 
dependent Dirichlet  and Pitman-Yor processes. These nonparametric 
models have been shown to be more flexible and robust, when
 compared to existing methods, for estimating the time-varying 
 number of objects, providing object labeling and 
identifying measurement to object associations.
Monte Carlo sampling methods are then proposed
to  efficiently learn trajectory of objects from noisy measurements. 
Using simulations, we demonstrate the 
 estimation performance advantage 
of the new methods when compared to existing 
algorithms such as the generalized labeled multi-Bernoulli filter.
\end{abstract}

\begin{keywords}
Bayesian nonparametric models, dependent Dirichlet process, dependent Pitman-Yor process, multiple object tracking, Markov chain Monte Carlo sampling
\end{keywords}

\section{Introduction}

Multiple object tracking under time-varying (TV) conditions
is a challenging and computationally intensive problem \cite{bar1990,Mah07,Wan17}.
It entails the  estimation of a TV  and unknown number 
of  objects at each time step, using measurements with no known
association to the objects.  Some recent methods to solving this problem
involve random finite set (RFS) theory methods, together with  
probability hypothesis density filtering and multi-Bernoulli filtering
 \cite{vo2014labled,Reu14,Vo2015,vo2017}.
  Many methods cluster objects to their associated estimated state
 after tracking. The labeled multi-Bernoulli filter, on the other hand, uses labeled RFS
to estimate the objects identity, although at a high computational cost and 
requiring high signal-to-noise ratios \cite{vo2014labled,Reu14}.
In \cite{Aok16}, maximum a posteriori estimates of object label uncertainties are
integrated with a multiple hypothesis tracking algorithm.  

In recent years,  the ubiquitous influence of Bayesian nonparametric
models has been well-established as a way to 
avoid the restrictions of parametric models. 
Infinite dimensional  random measures, 
such as the Dirichlet process (DP)  \cite{ferg1973,teh2011} and 
Pitman-Yor process (PYP) \cite{pitman1997,pitman2002} have
 replaced  finite mixture models for clustering, estimation and inference 
 \cite{antoniak1974,pitman2002,pitman1997}.
P{\'o}lya urn  approaches for TV  DP and PYP 
 mixtures  have been used as priors on parameters over observations. Note,
  however, that they do not capture the full dependency in a problem 
  and their marginal distributions are not well-defined  
  \cite{Ahmed2008,blei2011,caron2012,caron2017, moraffah2019use, moraffah2019tracking, moraffah2019nonparametric}.
 A process that does describe dependency among collections 
 of stochastic processes is the dependent DP  and 
 mixture models that can be used for clustering batch-sequential 
 data with a TV number of clusters  \cite{Mac99,Mac2000,campbell2013,neiswanger2014,topkaya2013, moraffah2019random, moraffah2018dependent, moraffah2019bayesian}.
 
In this paper, we propose a family  of prior distributions  
for time-dependent tracking algorithms 
constructed using the dependent DP  and PYP.
A Markov chain Monte Carlo (MCMC) inferential method integrates
 the distributions to update the time dependent states.  
 Our proposed methods  outperform existing ones and are 
 computationally efficient.  We also show that they have well-defined 
marginal distributions and hence provide an efficient way to 
perform inference. 
 They can  capture the full time-dependency
among object states and   can represent objects entering and
leaving a scene,  label object states and accurately estimate object 
cardinality  and trajectory.  They are also  simple to integrate
 with MCMC sampling methods for robust inference.

The rest of the paper is organized as follows.
Section \ref{track} summarizes the TV  tracking formulation
and Section \ref{processes} reviews the DP and PYP. 
In Section \ref{all_DDP}, we present the dependent DP  
 prior,  develop an MCMC inference method, and 
discuss algorithm convergence and properties.
 The extension to the dependent PYP as a prior 
 and its advantages over the DP 
are provided in Section \ref{all_DPYP}. 
Using simulations in Section \ref{sim}, we
demonstrate the methods' improved performance 
 when  compared 
 the RFS-based label multi-Bernoulli filter. 

\section{Multiple Object Tracking Formulation}
\label{track}

We consider the problem of tracking an unknown
and TV number of objects $N_k$ at  time step $k$.
The unknown state vector of the $\ell$th object, $\ell\eqq 1, \ldots, N_k$,
 is denoted by $\bx_{\ell, k}$. If the $\ell$th object is present 
 at time step $(k-1)$ and transitions to time step $k$,
 then  the object state  follows 
 $\bx_{\ell, k}\eqq 
 F(\bx_{\ell, k-1}) + \bu_{\ell, k-1}$,
 where $F(\cdot)$ is a transition function
 and $\bu_{\ell, k-1}$  is a modeling error random process. 
 The problem is further complicated by 
 the TV number of measurements $M_k$
 and the unknown object association to a given 
 measurement vector $\bz_{m,k}$, $m\eqq 1, \ldots, M_k$. 
 We assume that each measurement is generated by only one object
 and that the measurements are independent of one another. 
 If it is determined that the $m$th measurement originated 
 from the $\ell$th object,  the measurement equation for tracking is given 
 by $\bz_{m, k} \eqq  R(\bx_{\ell, k}) + \bw_k$;
 here, $R(\cdot)$ is a physics-based model describing the 
 relation between  the measurement to object state, and $\bw_k$ is  
   the measurement noise random process.
  Due to the large number of unknowns in the problem,
 additional functionality must be introduced. 
 Whereas RFS theory was used in previous 
 studies, we incorporate nonparametric methods,
 as will be demonstrated.
\section{Nonparametric Random Processes as Priors}
\label{processes}

\no {\it Dirichlet Process.}  \ The DP model defines a prior on the space of 
probability distributions  \cite{ferg1973,teh2011}.  
A  DP, $G$$\sim$$\text{DP}(\alpha, H)$, on an infinite 
dimensional space with parameter set
 $\Theta$,  concentration parameter
$\alpha$, and base distribution $H$,  is defined as
  \be
G(\theta) = \sum_{\ell =1}^{\infty} \pi_\ell \, \delta(\theta_\ell-\theta),  \ 
 \theta_\ell \sim H,   \  \pi_j\sim \text{GEM}(\alpha)
\label{dp}
\ee
where   $\text{GEM}(\alpha)$ follows the stick
 breaking representation \cite{sethuraman1994}
\ben
\pi_\ell = V_\ell \prod_{l = 1}^{\ell-1}(1-V_l), \   \text{where}  \ 
V_\ell  \sim \text{Beta}(1, \alpha), \  \  j = 1,2, \ldots 
\een
and $\delta(\theta_\ell- \theta)\eqq 1$ 
if $\theta\eqq \theta_\ell$, $\theta_\ell \!\! \in\! \!\Theta$, and zero otherwise.
$G(\theta)$ is a discrete probability random measure 
 with probability one. 
The DP mixture  (DPM) model 
for  parameters $y_n$, $n\eqq 1, \ldots, N$, 
is $y_n  \!  \mid \!  \theta_n \!  \sim \!  f(\cdot \mid \theta_n)$, where
$\theta_n \! \mid  \!  G \!  \sim \! G$ and   
$G \!  \mid \!  \alpha, H  \!  \sim\!   \text{DP}(\alpha, H)$.
The  independent and 
 identically distributed parameters  are drawn from 
$F(\cdot) \eqq \int_{\theta} f(\cdot\! \mid \!\theta) \, d G(\theta)$,
where $f(\cdot $$\mid$$ \theta)$ is the density and
$G(\theta)$ is the mixing distribution drawn according to a DP in  \eqref{dp}.
 It can be shown that the expected number of clusters using the DP
model varies exponentially as $\alpha \log N$. \\

\no {\it Pitman-Yor Process.} \ 
The PYP forms a large class of distributions on random probability measures that
contains DPs; it is a subclass of $d$-Gibbs 
processes and shares the DP essential properties.
Similar to the DP, it defines a prior on the space of probability distributions
over the infinite dimension space of parameters.  
A PYP,  $G$$\sim$$\text{PYP}(d, \alpha, H)$,  has discount parameter
  $d$, $0 \! \leq \! d \! < \! 1$,  concentration parameter $\alpha$, $\alpha \! > \!  -d$, 
  and base  distribution $H$.
  When  $d\eqq0$, the PYP simplifies to  $\text{DP}(\alpha, H)$.  
 A PYP realization  is a discrete random measure that can be constructed
using the stick breaking representation 
\be
G(\theta) = \sum_{\ell=1}^{\infty} \pi^*_\ell \,  \delta(\theta_\ell-\theta), \   \theta_\ell\sim H
\label{PY}
\ee
Here, $\pi^*_\ell$ is the size-biased order of $\pi_\ell$, 
 $\pi_\ell \eqq V_\ell \prod_{l =1}^{\ell-1}(1-V_l)$,
  and $V_\ell \sim \text{Beta}(1-d, \alpha+\ell \, d)$. Similar to the DPM,
the PYP mixture model for $y_n$, $n \eqq 1, \ldots, N$, is 
$y_n \! \mid \!  \theta_n \! \sim \!  f( \cdot \mid \theta_n )$, 
$\theta_n \!  \mid  \!  G \!  \sim G$, and
$G \!  \mid \!  \alpha, H \! \sim \!  \text{PYP}(d, \alpha, H)$.
The PYP expected number of clusters  follows the power-law 
$\alpha N^d$ \cite{teh2006}.

\section{Tracking with Dependent Dirichlet Process}
\label{all_DDP}

In order to capture the time-dependent evolution in 
 multiple object tracking, we exploit the dependent DP (DDP) \cite{Mac99}. 
The proposed DDP evolutionary Markov model (DDP-EMM)
approach is used to  learn  multiple object attributes 
over related information.  This information is based on 
dynamic dependencies in the state transition formulation.
In particular, the number of objects at time step $k$ 
depends on the number of objects  present
at the previous time step \kk  and on the objects popularity.
Also, the clustering index of  the $\ell$th
object state at time step $k$ depends on the clustering
index of the previous $(\ell\!-\!1)$ object states at the same time step. \\[-3mm]

\subsection{Construction of DDP Evolutionary Markov Model } 
\label{DDP_Model}

\begin{figure}[t]
\centering
\tikz \node [scale=0.55, inner sep=0] {
\begin{tikzpicture}
align = flush center,
\tikzstyle{plate caption} = [caption, node distance=0pt, 
inner sep=0pt, below left=3pt and 0pt of #1.south east]
\node[latent, shape=circle,draw, inner sep=1.5pt,
minimum size=3em] (thetak-1) {\Large $\bth_{\ell,k-1}$};%
\draw [->] (-3,0) -- (thetak-1);
\node[latent,right=3cm of thetak-1, shape=circle, draw, inner sep=1pt, 
minimum size=3em] (theta_tran) {\Large $\bth_{\ell, k \mid k-1}$ };
\node[latent,right=5.5 cm of thetak-1, shape=circle,
draw, inner sep=4.5pt, minimum size=3em] (theta_knew)
{\Large $\bth_{\ell, k}$};
\node[latent,right=3cm of theta_knew, shape=circle,
draw, inner sep=2pt, minimum size=3em, yshift=0.4 cm] (theta_k) 
{\Large $\bth_{\ell, k+1}$};
\edge{thetak-1}{theta_tran}
\draw [->,black] (theta_tran.north east) to [out=28, in=160] (theta_k.west);
\draw [->] (theta_k.east) -- (12.99,0.4);
\edge{theta_knew}{theta_k}
 \plate[inner sep=.25cm,xshift = -0.1cm, yshift=.25cm] {plate1} 
 {(thetak-1)} {\Large $\ell \eqq 1,\ldots, N_{k-1}$}; 
 \plate[inner sep=.25cm,xshift = -0.1cm, yshift=.25cm] {plate2} {(theta_tran)} 
 {\Large $\ell \eqq 1,\ldots, D_{k \mid k-1}$}; 
  \plate[inner sep=.25cm,xshift = -0.1cm, yshift=0cm] {plate3} 
  {(theta_knew)} {\Large $\infty$}; 
  \plate[inner sep=.25cm,xshift = -0.1cm, yshift=0cm] {plate4} {(plate2) (plate3)} {}; 
   \plate[inner sep=.25cm,xshift = -0.1cm,  yshift=0.25cm] {plate5} {(theta_k)}
   {\Large $\ell \eqq 1,\ldots, D_{k+1 \mid k}$}; 
\node[latent, below=1.5cm of thetak-1, xshift=1.5cm,
shape=circle, draw, inner sep=1.5pt, minimum size=3em]
(sk-1) { \Large ${\cal X}_{N_{k-1}, k-1}$};%
\node[latent, right = of sk-1, xshift = 2.5cm, shape=circle, draw, inner sep=-4pt, 
minimum size=5em] (sk) {\Large  ${\cal X}_{D_{k \mid k-1}, k}$};
\node[obs, below=1.5 cm of sk-1, shape=circle,draw,
inner sep=1.5pt, minimum size=3.05em] (zk-1) {\Large ${\cal Z}_{k-1}$};%
\node[obs, right = of zk-1, xshift = 2.5cm,
shape=circle,draw, inner sep=-4pt, minimum size=3.1em] (zk) {\Large ${\cal Z}_k$};%
\edge{sk-1}{sk}
\edge{sk-1}{zk-1}
\edge{sk}{zk}
\draw [->] (-1.2,-2.6) -- (sk-1);
\draw [dotted, ultra thick] (-2.5,-2.6) -- (-2.2,-2.6);
\draw [->] (sk.east) -- (8.6,-2.6);
\draw [dotted, ultra thick] (9.7,-2.6) -- (10,-2.6);
\edge{thetak-1}{sk-1}
\draw [->,black] (thetak-1.south)  to [out=-100, in=-210] (zk-1.north west); 
\edge{theta_knew}{sk}
\draw [->,black] (theta_knew.south east) to [out=-80, in=30] (zk.north east);
\edge{theta_tran}{sk}
\draw [->,black] (theta_tran.south)  to [out=-100, in=-210] (zk.north west);
    \end{tikzpicture}};
    \caption{Graphical model capturing temporal dependence.}
    \label{alg_gm}
\end{figure}
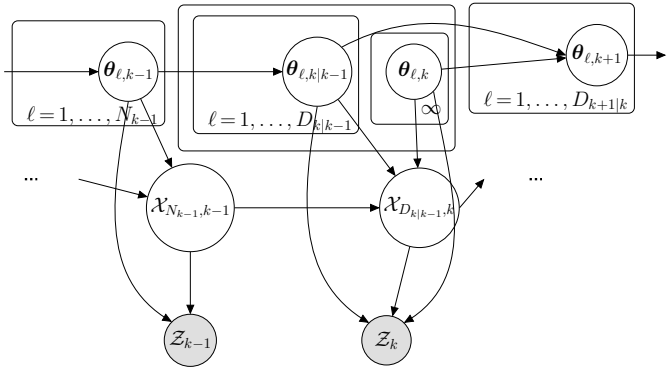

The DDP based construction of the  prior distributions of the 
multiple object states follows  three time evolution steps.
Specifically,  information from  time step \kks and 
from transitioning between  time steps  \kks and $k$  is used to predict
 information at time step $k$.   To aid in the construction and 
 dynamically track relevant assignments
 in relation  to the $\ell$th object and  $l$th cluster at time step $k$,
  we  define three indicators:  
cluster assignment  $c_{l, k}$, object transitioning   $\ts_{\ell , k}$,  and
cluster transitioning  $\lambda_{l, k}$.
    The three-step construction 
 is detailed  next and summarized  in Algorithm \ref{DDP_alg1}, and 
 a graphical model of the  evolutionary dependence 
 is shown in Figure \ref{alg_gm}.

 \vspace*{.1in}
\no {\bf Step 1.~ Previous time step:}

  At time step \kk, 
the  parameter sets 
\begin{equation}
{\cal X}_{N_{k-1}, k-1} =  \{ \bx_{1, k-1}, \ldots, \bx_{N_{k-1}, k-1} \}\notag
\end{equation}
 and 
 \begin{equation}
 \Theta_{N_{k-1}, k-1} = \{\bth_{1, k-1}, \ldots, \bth_{N_{k-1}, k-1} \}\notag
 \end{equation}
  are assumed available. Here, 
   the vectors $\bx_{\ell, k-1}$  and $\bth_{\ell, k-1}$ 
   denote the  $\ell$th object state and 
  the cluster parameter associated with
 the $\ell$th object,  $\ell \eqq 1,  \ldots, N_{k-1}$, respectively.
  To reduce computation, only 
 $D_{k-1}\!\! \leq \!\! N_{k-1}$  non-empty clusters are considered with 
 unique parameters 
 $\Theta^\star_{D_{k-1}, k-1} \!\! \subseteq \! \!\Theta_{N_{k-1}, k-1}$.
 It is also assumed that the  cardinality of the $l$th cluster
 (number of objects in the cluster) is 
 $v^\star_{l, k-1}$ and its  induced cluster assignment
  indicator  is $c_{l, k-1} \! \in\! \{1,\ldots, D_{k-1} \}$. The set of 
  induced cluster assignment indicators is denoted by  
$\ccC_{D_{k-1}, k-1}  \eqq \{ c_{1, k-1},  \ldots,  c_{ D_{k-1}, k-1} \}$.
  \vspace*{.1in}

\no {\bf Step 2.~ Transitioning between time steps.} 

From time steps  \kks to $k$,  the binary 
 object transitioning indicator $\ts_{\ell , k \mid k-1}$  
 determines the survival of the $\ell$th object.
The indicators follow a  Bernoulli process,  
with parameter $\text{P}_{\ell, k\mid k-1}$, 
associated with object transitioning.
If $\ts_{\ell , k \mid k-1}\eqq 1$, the object with 
state $\bx_{\ell, k-1}$ transitions to time $k$
with probability $\text{P}_{\ell, k\mid k-1}$
according to the Markov transition kernel
  $F_{\bth}(\bx_{\ell, k-1},  \cdot)$.
If $\ts_{\ell , k \mid k-1}\eqq 0$,  
the object leaves the scene with probability $(1-\text{P}_{\ell, k\mid k-1})$.

 An empty cluster is assumed to no longer  exist. 
Thus,  the binary cluster transitioning indicator $\lambda_{l, k\mid k-1}$ is defined
to keep track of the number of transitioning clusters $D_{k\mid k-1}$.
Specifically, if the $l$th cluster cardinality  satisfies $v^\star_{l,  k-1}\!\! \geq \!\!  1$,  
 $\lambda_{l, k\mid k-1}\eqq 1$; otherwise,
$\lambda_{l, k\mid k-1}\eqq 0$, $l\eqq 1, \ldots, D_{k-1}$
and  $D_{k\mid k-1} \eqq  \sum_{l=1}^{D_{k-1}} \lambda_{l, k\mid k-1}$.
The  $l$th transitioning cluster cardinality is set to $v^\star_{l, k \mid k-1}$,
and the parameter   of the transitioning
cluster  associated with the $\ell$th  object is 
$\bth^\star_{\ell, k \mid k-1}$.

 \vspace*{.1in}
\no {\bf Step 3.~ Current time step.}

The three cases discussed next are used to
 formulate the distributions of both the
DP cluster parameter $\bth_{\ell, k}$ and its associated object state $\bx_{\ell, k}$ 
at time step $k$. The number of non-empty clusters is 
set to $D_k\eqq D_{k\mid k-1}$, and  the cardinality of 
the $l$th transitioning cluster  is set to 
$v_{l, k} \eqq v^\star_{l, k \mid k-1}$, $l\eqq 1, \ldots, D_k$. \\[-3mm]

\no  Case D1:  
The $\ell$th transitioned object, with 
$\ts_{\ell , k \mid k-1}\eqq 1$,  is assigned to a transitioned cluster already 
  occupied by at least one of the  $(\ell \!-\!1)$  previous objects. 
  As the cluster assignment set $\ccC_{D_k, k}$ at time step $k$
  induces an  infinite exchangeable random partition, 
  it is assumed that the $\ell$th  object is the last  to be clustered 
  and  selects one of these clusters with  probability  
\be
& & \hspace*{-.3in} 
\Pi^{(1)}_{k} = \text{Pr}\Big (\text{select} \  l \text{th} \ \text{cluster}, 
l \! \leq \! D_k \mid  \Theta_{\ell-1, k} \Big )  \nonumber \\[-2mm]
& & \hspace*{-.3in} = \dfrac{1}{g_k}\Big (  v_{l, k} + 
  \sum_{j=1}^{D_{k-1}}
v^\star_{j, k \mid k-1} \, \lambda_{j, k \mid k-1} \,  
 \delta(c_{j, k} - c_{l, k}) \Big ),  \label{case1}
\ee
 $g_k \eqq (\ell-1) +  \alpha 
 + \sum_{l=1}^{\ell-1} \sum_{j=1}^{D_{k-1}} 
v^\star_{j, k \mid k-1}   \lambda_{j, k \mid k-1} \delta(c_{j, k} - c_{l, k} )$ 
and  $\alpha$$>$$0$ is the concentration parameter. 
The cluster parameter $\bth_{\ell, k}$  associated to the $\ell$th object
is drawn from the transitioning kernel $\varphi(\bth^*_{\ell, k-1},  \bth_{\ell, k} )$.
With probability $\Pi^{(1)}_k$, the
$\ell$th object state prior distribution is drawn as 
\be
 p_1\Big (\bx_{\ell, k} \mid \Upsilon_{\ell, k}  \Big ) 
 = F_{\bth}( \bx_{\ell, k-1}, \bx_{\ell, k}) \; f( \bx_{\ell, k} \mid \bth^*_{\ell, k}),
\label{state1}
\ee
where  $F_{\bth}(\bx_{\ell, k-1}, \cdot)$ is the object transitioning kernel
and $f(\bx_{\ell, k} \!\mid \! \! \cdot)$ is the object state density.
The conditional parameter set  in \eqref{state1} is 
 $\Upsilon_{\ell, k} \eqq \{ {\cal X}_{\ell-1, k}, {\cal X}_{\ell, k\mid k-1}, 
 \Theta_{\ell, k}, \Theta^*_{\ell, k \mid k-1} \}$. \\[-2mm]

\no Case D2:  The $\ell$th object is assigned to
one of the transitioning clusters 
that has not yet been assigned to any of the  $(\ell-1)$ previous objects.  
The object selects one of these clusters  with probability 
\be
& & \hspace*{-.3in} 
\Pi^{(2)}_k = \text{Pr}\big (\text{select} \  l \text{th} \ \text{cluster}, 
 \ l \leq D_k \,  \mid   \Theta_{\ell-1, k} \big )  \nonumber \\[-1mm]
& & \hspace*{.1in} =  \dfrac{1}{g_k} \Big ( 
  \sum_{j=1}^{D_{k-1}} \! \! 
v^\star_{j, k \mid k-1} \lambda_{j, k \mid k-1}   \delta( c_{j, k} - c_{l, k}) \Big )\, .
\label{case2}
\ee
The cluster parameter $\bth_{\ell, k}$  associated to the $\ell$th object
is drawn from $\varphi(\bth^*_{\ell, k-1},  \bth_{\ell, k} )$. With 
probability $\Pi^{(2)}_k$, the state prior  distribution is   drawn as 
\be
 & & \hspace*{-.3in} p_2(\bx_{\ell, k} \mid  \Upsilon_{\ell,  k})  =  
 F_{\bth}( \bx_{\ell, k-1}, \bx_{\ell, k}) \, f( \bx_{\ell, k} \mid \bth^*_{\ell, k}) \nonumber \\
 & & \hspace*{1in} \cdot \, \varphi(\bth^*_{\ell, k-1},  \bth_{\ell, k-1})\, .
 \label{state2}
\ee

\no Case D3:  The $\ell$th object does not belong to an
 existing cluster, and  a  new  cluster  is formed 
 with associated parameter vector 
 drawn from the base of DP$(\alpha, H)$, $\bth_{\ell, k} \sim H$.  The object is selects this cluster with   probability
\be
\Pi^{(3)}_k=  \text{Pr}\big (\text{new cluster} \mid  \Theta_{\ell-1, k}  \big )  =  \alpha/g_k \, .
\label{case3}
\ee
The state prior distribution is drawn as 
\be
p_3(\bx_{\ell, k} \mid  \Upsilon_{\ell, k}) 
 =  \int_{\bth} 
 f( \bx_{\ell, k} \mid \bth) \, d H( \bth)\, .
 \label{state3}
\ee

Assuming  the space of state parameters is
Polish,  the DDP in Cases D1--D3  define marginal DPs at each time
step $k$, given the DDP configurations at time step $(k-1)$.
Specifically, 
\be
\text{DDP-EMM}_k | \text{DDP-EMM}_{k-1} \sim 
\text{DP} \Big( \alpha, H_0 \Big )
\label{DDP-EMM}
\ee
with base distribution  
\be
& & \hspace*{-.2in} H_0=  
\sum_{\substack{j=1  \\[0.2mm] \bth_{j, k} \in \Theta_{D_k, k}}}^{D_k}
 \hspace*{-3mm} \Pi^{(1)}_k \,  
\delta( \bth_{j,k} - \bth_{\ell, k})  +  \Pi^{(3)}_k \, H(\bth_{\ell, k}) \nonumber  \\
  & & \hspace*{-.3in}  +  \sum_{\substack{j=1 \\[1mm]
  \bth^\star_{j, k} \in  \Theta^\star_{D_k, k \mid k-1}  \setminus \Theta_{D_k, k}} }^{D_k} 
\hspace*{-8mm}
 \Pi^{(2)}_k\,  \varphi(\bth^\star_{\ell, k-1}, \bth_{\ell, k} ) \,  
 \delta( \bth_{j, k} - \bth_{l, k}).   
 \label{HH}
\ee
This  model  also allows the variation and labeling of 
clusters  as it is defined in the space of partitions.
\begin{algorithm}[t] 
\begin{algorithmic}
\caption{DDP-EMM construction of prior distribution}
\label{DDP_alg1}
\STATE 
\hspace*{-.25in} Step 1. {\bf At time} \kk, parameters available from Step 1
\STATE \blll Object state parameters set, ${\cal X}_{N_{k-1}, k-1}$  
\STATE \blll Cluster parameters set, $\Theta_{N_{k-1}, k-1}$ 
\STATE \blll Non-empty cluster unique parameters set, $\Theta^\star_{D_{k-1}, k-1}$ 
 \STATE \blll Cluster assignment indicator set, $\ccC_{D_{k-1}, k-1}$
 \STATE \blll Cardinality of $l$th non-empty cluster,  $v^\star_{l, k-1}$
\STATE \vspace*{-3mm}
\STATE \hspace*{-.2in}  Step 2.~{\bf Transitioning} between time steps $(k$$-$$1)$ and $k$ 
%
\STATE  \hspace*{-.18in} Draw 
$\ts_{\ell , k \mid k-1}$$\sim$$\text{Bernoulli} (\text{P}_{\ell, k\mid k-1})$,
$\ell$th object transitioning indicator
\STATE \hspace*{-.18in}  If $\ts_{\ell , k \mid k-1}\eqq 1$,  $\ell$th object
survives with probability  (w.pr.)  $\text{P}_{\ell, k \mid k-1}$  and
transitions with kernel $F_{\bth}( \bx_{\ell, k-1},  \bx_{\ell, k})$
\STATE \hspace*{-.18in}  If  $l$th 
cluster cardinality satisfies $v^\star_{l, k-1} \geq 1$, set 
$l$th cluster transitioning indicator  to $\lambda_{l, k \mid k-1}\eqq 1$,
$l\eqq 1, \ldots, D_{k-1}$
\STATE  \hspace*{-.18in} Set {\footnotesize $\#$} of transitioning clusters to 
$D_{k\mid k-1} \eqq \sum_{l} \lambda_{l, k\mid k-1}$
\STATE  \hspace*{-.18in} Denote  cardinality of  $l$th transitioning  cluster
by $v^\star_{l, k \mid k-1}$ 
\STATE  \hspace*{-.18in}  Denote parameter vector of transitioning cluster associated
with $\ell$th object by $\bth^\star_{\ell, k \mid k-1}$
\STATE \vspace*{-3mm}
  \STATE \hspace*{-.2in} Step 3.~{\bf At time } $k$
 \STATE \hspace*{-.14in} Set $D_k \eqq D_{k \mid k-1}$ 
 \FOR{$\ell \eqq 1$ \TO $D_k$}
\STATE \hspace*{-.09in} Set $v_{\ell, k}  \eqq v^\star_{\ell, k\mid k-1}$ 
\IF{$\ell \leq D_k$ \AND $\ell$th cluster already selected}
\STATE \hspace*{-4mm}  Draw $\bth_{\ell, k}\sim \varphi(\bth^*_{\ell, k-1},  \bth_{\ell, k})$ with 
\STATE \hspace*{-4mm}  Draw  
$\bx_{\ell, k}$$\mid$$\bth_{\ell,k}\!\sim\! p_1(\bx_{\ell, k} \mid \! \cdot)$
 in \eqref{state1} with prob.~$\Pi^{(1)}_k$ in \eqref{case1}
\ELSIF{$\ell \leq D_k $ \AND $\ell$th cluster not yet selected}
\STATE \hspace*{-4mm} Draw $\bth_{\ell, k}\sim \varphi(\bth^*_{\ell, k-1},  \bth_{\ell, k})$ 
\STATE \hspace*{-4mm} Draw  $\bx_{\ell, k}$$\mid$$\bth_{\ell,k}\!\sim\! 
p_2(\bx_{\ell, k} \mid \! \cdot)$
 in \eqref{state2} with prob.~$\Pi^{(2)}_k$ in \eqref{case2}
\ELSE
\STATE  \hspace*{-4mm} Draw $\bth_{\ell, k} \sim H$ for new cluster associated with $\ell$th object
\STATE \hspace*{-4mm} Draw 
$\bx_{\ell, k}$$\mid$$\bth_{\ell, k}\!\sim\! p_3(\bx_{\ell, k} \mid \! \cdot)$ 
in \eqref{state3} with prob.~$\Pi^{(3)}_k$ in \eqref{case3}
\ENDIF
 \ENDFOR 
\RETURN{ $\{ \bx_{1, k}, \bx_{2, k}, \ldots, \ldots \}$, 
$\{\bth_{1, k}, \bth_{2, k}, \ldots,  \ldots \}$ \\}
\end{algorithmic}
\end{algorithm}

\subsection{Learning Model}
\label{ddplearning}

Considering  the measurements  $\ccZ_k \eqq \{ \bz_{1, k}, \ldots, \bz_{M_k, k} \}$ 
 at time step $k$, the modeled prior distribution 
is used with an MCMC method to perform  inference.
 The measurements  are assumed  independent,  
 each  generated from one object, and  unordered;  the 
$m$th measurement is not necessarily associated to
the $\ell$th object,  $m\neq \ell$. 
The posterior distribution is used to estimate the objects states
and find the time-dependent object cardinality. 
 As the DDP is used to label the object states at time
step $k$,  DPMs are used to learn and assign each measurement to its
associated object  identity.   The mixing
measure is drawn from  the  DDP in Algorithm \ref{DDP_alg1}
 to  infer the likelihood distribution
\be
 \bz_{m, k} \mid  \bth_{\ell, k}, \bx_{\ell,k} \sim R(\bz_{m,k} \mid \bth_{\ell,k}, \bx_{\ell,k} ) 
\label{likelihood} 
\ee
where $R(\bz_{m,k} \mid \bth_{\ell, k}, \bx_{\ell, k})$  is a distribution that 
depends on
the measurement likelihood function.  Algorithm
\ref{alg2} summarizes the mixing process that 
associates measurements to objects.

The Bayesian posterior to estimate the target trajectories is
efficiently implemented using a  Gibbs sampler inference scheme.
The scheme iterates  between sampling the object 
states and the dynamic DDP parameters, and it is based on 
 the discreetness of the DDP  \cite{Mac2000,Mac1998}. 
 The Bayesian posterior is  
 \be
p( \bx_{\ell, k} \mid \ccZ_k ) = \int P(\bx_{\ell, k} \mid \ccZ_k,  \Theta_{D_k, k} ) \, 
d G(\Theta_{D_k, k} \mid \ccZ_k)
\label{predictive}
\ee
where $G(\Theta_{D_k, k} \!\! \mid \!\! \ccZ_k)$ is the cluster parameter posterior distribution
given  the measurements.  As direct computation of \eqref{predictive} is 
intensive \cite{antoniak1974}, Gibbs sampling is used 
to predict  $\bx_{\ell, k}$ using 
 $p(\bx_{\ell, k} \! \! \mid \! \!  \ccZ_k,  \Theta_{D_k, k}) \eqq 
 p(\bx_{\ell, k}\!  \mid \!  \Theta_{D_k, k})$, which is evaluated as 
$p(\bx_{\ell, k}\!  \mid \!  \Theta_{D_k, k})\eqq 
\int p(\bx_{\ell, k} \! \mid \! \bth_{\ell, k})  \, d \pi(\bth_{\ell, k} \!  \mid \!  \Theta_{D_k, k})$.
Here,  the posterior  of $\bth_{\ell, k}$ given the remaining parameters is
\be
& & \hspace*{-.4in} \pi(\bth_{\ell, k} \mid \Theta_{D_k, k} )  = \hspace*{-3mm} 
\sum_{\substack{j=1,  j\neq \ell \\[1mm] \bth_{j, k} \in \Theta_{D_k, k} }}^{D_k}
\hspace*{-3mm}  \Pi^{(1)}_k \,  \delta( \bth_{\ell,k} - \bth_{j,k}) + 
 \Pi^{(3)}_k\,  H(\bth_{\ell, k}) \nonumber  \\
 & & \hspace*{-.2in} 
 + \hspace*{-.2in} \sum_{\substack{j=1,  j\neq \ell  \\[1mm]
  \bth_{j, k} \in  \Theta^\star_{D_{k \mid k-1}, k \mid k-1}  
  \setminus \Theta_{D_k, k}}}^{D_{k \mid k-1}}  \hspace*{-.2in}
  \!\!\! \Pi^{(2)}_k\,  \varphi(\bth^\star_{\ell, k-1}, \bth_{\ell, k} ) 
  \,  \delta( \bth_{j, k} - \bth_{\ell, k}).  
\label{posttheta}
\ee
The distribution obtained by combining the prior 
  with the likelihood is 
\be
& & \hspace*{-.4in} \bth_{\ell,k} \mid \Theta^{(\ell)}_k,  \ccZ_k  \ \sim   \
 \sum_{j=1}^{D_k} \xi_{j, k} \,  \delta(\bth_{\ell,k}- \bth_{j,k})
 \nonumber  \\
 & & \hspace*{-.2in}
+  \sum_{\substack{ j=1 \\  j \notin \ccC_{D_k, k} }}^{D_{k\mid k-1}}
\beta_{j, k}  \, R_{j, k}(\bth_{\ell,k}) 
+  \gamma_{\ell,k}\, H_{\ell}(\bth_{\ell,k}),
\label{Gibbs}
\ee
where 
$\Theta^{(\ell)}_k\eqq \{ \bth_{1, k}, \bth_{2, k}, \ \ldots, \ \bth_{\ell-1, k}, \bth_{\ell+1, k},  \hfill
\ldots, \bth_{D_{k\mid k-1}, k} \}$,   
$R_{j, k}(\bth_{\ell,k}) \eqq R(\bz_{\ell,k} \mid \bx_{j, k}, \bth_{j, k})$, 
\ben
& & \hspace*{-.3in}
 \xi_{j, k} = 
\frac{R_{j, k}(\bth_{\ell,k}) }{g_k} 
\Big ( v_{j, k} +  \!\! \displaystyle \sum_{i=1}^{D_{k|k-1}}  \!\!
v^\star_{i, k \mid k-1}  \lambda_{i, k \mid k-1}   
\delta(c_{i, k} - c_{j, k})\Big )  \\
& & \hspace*{-.1in} \beta_{j,k} = \frac{1}{g_k} 
\sum_{\substack{ i=1 \\  i \notin \ccC_{D_k, k} }}^{D_{k\mid k-1}}
v^\star_{j, k \mid k-1} \lambda_{j, k\mid k-1} \\
& & \gamma_{\ell, k} = 1 -  \sum_{j=1}^{D_k } \zeta_{j, k} -  \sum_{\substack{ j=1 \\  j \notin \ccC_{D_k, k} }}^{D_{k\mid k-1} } \beta_{j, k}.
\een
Also,  $dH_{\ell}(\bth_{\ell, k}) \! \propto \!
R_{j, k}(\bth_{\ell,k})  \, dH(\bth_{\ell, k})$, 
 $H$ is the base distribution on $\bth_{\ell, k}$, and 
  $g_k$ is defined below \eqref{case1}. 
The derivation  is provided in  \cite{Bahman_rept}.

\begin{algorithm}[t]
\caption{Infinite mixture model for measurement-to-object association}
\begin{algorithmic}
\STATE {\bf Input}: $\{ \bz_{1, k},   \ldots, \bz_{M_k, k}\}$, measurements
\STATE From construction of prior distribution from Algorithm \ref{alg_gm}
\STATE {\bf Input}: Object state vectors $\{ \bx_{1, k},  \bx_{2, k}, \ldots \}$
 \STATE {\bf Input}: Cluster parameter vectors   $\{ \bth_{1, k},  \bth_{2, k}, \ldots \}$
  \STATE {\bf Input}: Cluster assignments 
  ${\cal C}{\cal A}_{k-1}\eqq  \{ \ccC_{D_k, 1}, \ldots, \ccC_{D_k, k-1} \}$
\STATE { \bf At time} $k$:
\FOR{$ m\eqq 1$ \TO $M_k$}
 \STATE Draw $\bz_{m,k} \mid  \bx_{\ell, k}, \bth_{\ell, k}$ from 
 Equation \eqref{likelihood} 
   \RETURN  $\cca_{D_k,k}$, induced cluster assignment indicators   
      \ENDFOR
      \STATE  {\bf Update}:  
      ${\cal C}{\cal A}_k\eqq  {\cal C}{\cal A}_{k-1}  \cup \cca_{N_k}$
\RETURN  $D_k$ (number of clusters) and  ${\cal C}{\cal A}_k$
\RETURN  posterior of $\bz_{m, k}  \mid  \bx_{\ell, k}, \bth_{\ell, k}$, 
 $m\eqq 1, \ldots, M_k$
\end{algorithmic}
\label{alg2}
\end{algorithm}

\subsection{DDP-EMM Approach Properties}
\label{prop}

\paragraph{Convergence}

In the Gibbs sampler,  it can be shown that the  transition kernel converges to the 
posterior distribution   for almost all initial conditions. 
If after $n$ iterations of the algorithm, $\Psi^{(n)}_k(\bth_{\ell, 0}, \Theta_{D_k, k})$ 
is the transition kernel for the Markov chain starting at $\bth_{\ell, 0}$ 
and stopping in  the set $\Theta_{D_k, k}$,   
then it can be shown to converge to the 
posterior  $G(\Theta_{D_k, k} \! \mid \! \ccZ_k)$
 given the measurements $\ccZ_k$ at time step $k$. Specifically, 
\ben
|| \Psi^{(n)}_k(\bth_{\ell, 0}, \cdot) - 
G(\cdot \! \mid \! \ccZ_k)  ||_{\text{TVN}}
\rightarrow 0 \ \text{as} \  n\rightarrow \infty, 
\een
 for almost all initial conditions $\bth_{\ell, 0}$  in the total variation norm (TVN)
 (see \cite{escobar1995,escobar1994} in relation to the Gaussian distribution).
 The proof of convergence can be found in \cite{Bahman_rept}.

\paragraph{Exchangeability}
The infinite  exchangeable random partition induced by 
$\ccC_{D_k, k}$ at time $k$ follows  the 
exchangeable partition probability function (EPPF) \cite{aldous1985}
\be
p( \ccV^\star_{D_k, k}) = 
\frac{\alpha^{D_k}}{\alpha^{\left[N_k\right]}} 
\prod_{j=1}^{D_k} ( v^\star_{j,k} - 1)
\label{EPPFDir}
\ee
where $\ccV^\star_{D_k, k} \eqq \{  v^\star_{1,k},  \    \ldots,  \ v^\star_{D_k, k} \}$, 
$v^\star_{j,k}$ is the cardinality of  the  cluster 
with assignment indicator $c_{j, k}\in \ccC_{D_k, k}$, 
$\alpha^{\left[n\right]}\eqq  \alpha (\alpha+1) \dots (\alpha+n-1)$, and 
 $D_k$ is the number of unique cluster parameters.
Due to the variability of $N_k$,  there is an important relationship
between the partitions based on $N_k-1$ and $N_k$.
In particular, the EPPF  in \eqref{EPPFDir}  
  based on the partitions on $N_k$ and $N_k-1$ objects, 
   given the configuration at time $(k-1)$, satisfies
\be
p_{N_k-1}( \ccV^\star_{D_k, k}) = 
 \sum_{j = 1}^{D_k}    p_{N_k}( \ccV^\star_{j, k}) 
 +   p_{N_k}\big ( \{ \ccV^\star_{D_k, k}, 1 \} \big )
 \label{partition}
\ee
where 
$\ccV^\star_{j, k} \eqq \{  v^\star_{1, k},  \    \ldots,  \  (v^\star_{j, k}+1),  
v^\star_{j+1, k}, \ldots, \ v^\star_{D_k, k} \}$. 
Equation \eqref{partition} entails a 
notion of consistency of the partitions in the 
distribution sense. The equation  holds 
due to the Markov property of the process 
given the configuration at time $(k-1)$. 

\paragraph{Consistency}

We consider $r_{\bth_0}$ to be the true density of observations with probability measure 
$R_{\bth_0}$.  Then, if $r_{\bth_0}$ is in the Kullback-Leibler
 (KL) support of the prior distribution
in the space of all parameters \cite{Choi08}, then  
the posterior distribution $G(\cdot  \! \mid \! \ccZ_k)$ 
can be shown to  be weakly consistent at $r_{\bth_0}$ 
 It is also important to investigate the  posterior contraction rate  as it is 
 highly related to posterior consistency.
This rate shows how fast the  posterior distribution 
approaches the true parameters from which the measurements are generated.  
As detailed in \cite{Bahman_rept},  the contraction rate matches the minimax rate 
for density estimators. Hence, the DDP prior constructed through the proposed model 
achieves the optimal frequentist rate.

\section{Tracking with Dependent Pitman-Yor Process}
\label{all_DPYP}

Whereas the expected number of unique clusters used by  
the DP model during transitioning is $\alpha \log{(N_k})$,
 the number used by the PYP model  follows the power law 
$\alpha N_{k}^d$. Here,  $N_k$ is the number 
of objects  to be clustered,   $\alpha$ is a concentration 
parameter and  $d$ is a discount parameter. 
With this additional parameter, the PYP model has a higher 
probability of having a large number of unique clusters   \cite{teh2006}.
 Also, clusters with only a small number of objects have a lower probability of
selecting new objects.  This more flexible model 
is  better matched to  tracking a TV number of objects as the 
larger number of available clusters  ensures all
dependencies are captured. The proposed  dependent 
PYP (DPYP)  state transitioning prior (DPYP-STP) method
 is presented next. 
  
\noindent {\bf Construction Model} \ The  construction of the DPYP prior distribution 
 follows  Steps 1 and 2 for the DDP 
in Section \ref{DDP_Model}.  As in Step 3 for the DDP,
it is assumed that there are $D_k\eqq D_{k\mid k-1}$
non-empty clusters at time step $k$,
and that the cardinality of the $l$th transitioning cluster  is set to 
$v^\star_{l, k \mid k-1}\eqq v_{l, k}$, $l\eqq 1, \ldots, D_k$.
The  state prior distributions are 
 drawn following Cases D1-D3.  However, the 
 probability of an object selecting a  particular cluster varies, as provided next 
in Cases P1--P3.\\[-2mm]

\no Case P1:   
The $\ell$th object  is assigned, with probability 
\be 
& & \hspace*{-0.3in} \Pi^{(1)}_ k =
\text{Pr}\Big (\text{select} \  l \text{th} \ \text{cluster}, 
l \! \leq \! D_k \mid  \Theta_{\ell-1, k} \Big )  \nonumber \\
 & & \hspace*{-.3in} 
= \dfrac{1}{g_k} \Big (  v_{\ell, k} - d + \!\!
  \sum_{j=1}^{D_{k-1}} \!\!\!
v^\star_{j, k \mid k-1} \, \lambda_{j, k \mid k-1} \,  
 \delta( c_{j, k} - c_{\ell, k}) \Big )  
 \label{eq1}
\ee
to a transitioning cluster that is already 
  occupied by at least one of the  $(\ell \!-\!1)$  previously transitioned objects. 
  In \eqref{eq1}, 
 $g_k \eqq  b_{\ell-1} +  \alpha  + \sum_{l=1}^{\ell-1} \sum_{j=1}^{D_{k-1}} 
v^\star_{j, k \mid k-1}   \lambda_{j, k \mid k-1} \delta(c_{j, k} - c_{l, k} )$,
$b_{\ell-1}\eqq  \sum_j^{\ell-1}  v_{j, k}$, 
 $\alpha\!\!>\!\! - d$ and  $0 \!\! \leq \!\! d \!\! < \!\! 1$.  \\[-2mm]
 
\no Case P2: 
The $\ell$th object selects a  transitioning cluster that is not yet occupied
by the previous $(\ell \!-\!1)$  objects   with probability
\be
& & \hspace*{-.3in} 
\Pi^{(2)}_k = 
\text{Pr}\big (\text{select} \  l\text{th} \ \text{cluster}, 
l \leq  D_k   \mid   \Theta_{\ell-1, k} \big )  \nonumber \\
& & \hspace*{-.1in}  = \dfrac{1}{g_k} \Big ( 
  \sum_{j=1}^{D_{k-1}} \! \! 
v^\star_{j, k \mid k-1} \lambda_{j, k \mid k-1}   \delta( c_{j, k} - c_{l, k}) \Big ) - d \, .
\label{eq2}
\ee

\no  Case P3: The $\ell$th object selects a new 
cluster with probability
\be
\Pi^{(3)}_k= \! \text{Pr} 
\big (\text{new cluster} \mid \Theta_{\ell-1, k} \big ) 
 = \frac{ d \, D_k^{(\ell-1)} +\alpha}{g_k}
\label{eq3}
\ee
where $D_k^{(\ell-1)}$ is the total number of clusters
at time step $k$ used by the previous $(\ell -1)$ objects.  \\[-2mm]

Note that, in Cases P1 and P2, the cluster parameter is drawn from 
a transitioning kernel.  In Case P3, 
the  cluster parameter is drawn from the base distribution of $\text{PYP}(d, \alpha, H)$ as
$\bth_{\ell, k} \sim H$.
If the state parameter space  is
separable and metrically topologically complete, 
 the DPYP in Cases P1--P3  define marginal PYPs 
 given the DPYP configurations at  time  step $(k-1)$.
 In particular, 
\be
\text{DPY-STP}_k \mid \text{DPY-STP}_{k-1} \sim 
\text{PYP}\Big(d, \alpha, H_1\Big)
\label{condd}
\ee
with base distribution defined as in \eqref{HH}, but the  with 
probabilities  given by \eqref{eq1}-\eqref{eq3}.

\noindent {\bf Learning Model} \ The DPYP  prior distribution is integrated with MCMC
 to perform inference, as in Section \ref{ddplearning}.
The major difference from the DDP-EMM approach is 
that the mixing measure is drawn from the DPYP.
DPMs are also used to learn the measurement to object associations. 
 Note that the DPYP-STP and DDP-EMM algorithms
  are closely related;  setting $d\eqq 0$ in the DPYP-STP simplifies to 
  the DDP-EMM.
  
\section{Simulation Results}
\label{sim}

We demonstrate the performance of the two proposed methods,
DDP-EEM  and DPY-STP,  that are  based on 
using dependent Bayesian nonparametric models to account for
dynamic dependencies in multiple object tracking.
Unless otherwise stated, the simulations assume the  following parameters.  
The coordinated turn model (CTM) (constant turn rate) is used 
as the motion model in tracking multiple objects moving 
in the two-dimensional (2-D) plane. Using CTM, 
the state vector of the $\ell$th object is 
$\bx_{\ell, k} \eqq [x_{\ell, k} \   \dot{x}_{\ell, k}\   y_{\ell, k}  \  \dot{y}_{\ell, k} \  
\omega_{\ell, k}]^T$, $\ell\eqq 1, \ldots, N_k$,
 where $(x_{\ell, k},  y_{\ell, k})$ and $(\dot{x}_{\ell, k}, \dot{y}_{\ell, k})$
 are the 2-D coordinates for position and velocity, 
 respectively, and  $\omega_{\ell, k}$ is the  turn rate.
 Using the state space model notation from Section \ref{track},
 if the  $\ell$th object transitions  between  time steps, 
 the transition equation is 
 $\bx_{\ell, k}\eqq F(\bx_{\ell, k-1}) + \bu_{\ell, k-1} \eqq
 D \bx_{\ell, k-1} +  \bu_{\ell, k-1}$,  where 
 \ben
 D = 
 \begin{bmatrix}
1 &  \dfrac{\sin(\omega_{k-1})}{\omega_{k-1}} & 0 
& - \dfrac{1-\cos(\omega_{k-1})}{\omega_{k-1}} & 0\\[3mm]
0  &\cos(\omega_{k-1}) & 0 &-\sin(\omega_{k-1}) & 0\\[2mm]
0 & \dfrac{1-\cos(\omega_{k-1})}{ \omega_{k-1}}  
& 1 & \dfrac{\sin(\omega_{k-1})}{\omega_{k-1}} & 0\\[3mm]
0 & \sin(\omega_{k-1}) & 0 & \cos(\omega_{k-1}) &0  \\[2mm]
0 & 0 & 0& 0& 1 \end{bmatrix},
 \een
 $\bu_{\ell, k-1}$  is zero-mean Gaussian  with covariance  matrix
 \ben
 Q_u = \begin{bmatrix}
  \sigma^2/4 &  \sigma^2/2& 0 & 0 & 0 \\
 \sigma^2/2 & \sigma^2 & 0 & 0 & 0 \\
 0 & 0 & \sigma^2/4 &  \sigma^2/2 & 0 \\
 0 & 0 & \sigma^2/2 & \sigma^2 & 0 \\
 0 & 0 &  0 & 0 &  \sigma_u^2
 \end{bmatrix} ,
 \een
 $\sigma\eqq15$ m/s$^2$ and  $\sigma_u \eqq \pi/180$ radians/s.
The  measurement  equation for the $\ell$th  object is
 $\bz_ k \eqq [\phi_k \ r_k]^T \eqq  R(\bx_{\ell, k}) + \bw_k$
  where $R(\bx_{\ell, k})\eqq [ \arctan{(y_{\ell, k}/x_{\ell, k})} \, 
 (x^2_{\ell, k} +  y^2_{\ell, k})^{1/2} ]^T$,  
 with bearing  $\phi_k \!\!\in  \!\!(-\pi/2,  \pi/2)$ and range 
  $r_k  \!\!\in  \!\!(0, 2,000)$ m. 
   The measurement noise $\bw_k$  is assumed zero-mean Gaussian with 
  covariance $Q_w\eqq \text{diag}(25, (\pi/180)^2)$.
  
  The cluster parameter prior distribution $H$ was generated 
using   a  normal-inverse Wishart distribution (NIW), 
 $\text{NIW}(\mu_0 , 0, \nu, {\cal I})$, with 
 identity matrix  ${\cal I}$,  $\mu_0\eqq 0.001$ and $\nu\eqq 50$.
The prior for the concentration parameter used 
The Gamma distribution $\Gamma(\alpha; 1, 0.1)$
was used as  the concentration parameter prior.
The object transitioning probability is fixed to 
$\text{P}_{\ell, k \mid k-1}\eqq 0.95$, for all $\ell$.  
All simulations use  100 times steps and 
10,000  Monte Carlo realizations; the signal-to-noise ratio (SNR) is -3 dB.
The proposed methods are compared with the generalized labeled multi-Bernoulli filter (GLMB) 
that models time-variation using  labeled RFS 
\cite{vo2014labled,Reu14,Vo2015,vo2017}.
The optimal sub-pattern assignment (OSPA) metric,  with order
$p\eqq1$  and cut-off $c\eqq 100$, is used to compare the methods. 
For this metric, lower the metric values indicate higher performance. \\

\begin{figure}[htb]
\begin{center}
\resizebox{2.7in}{1.6in}{\includegraphics{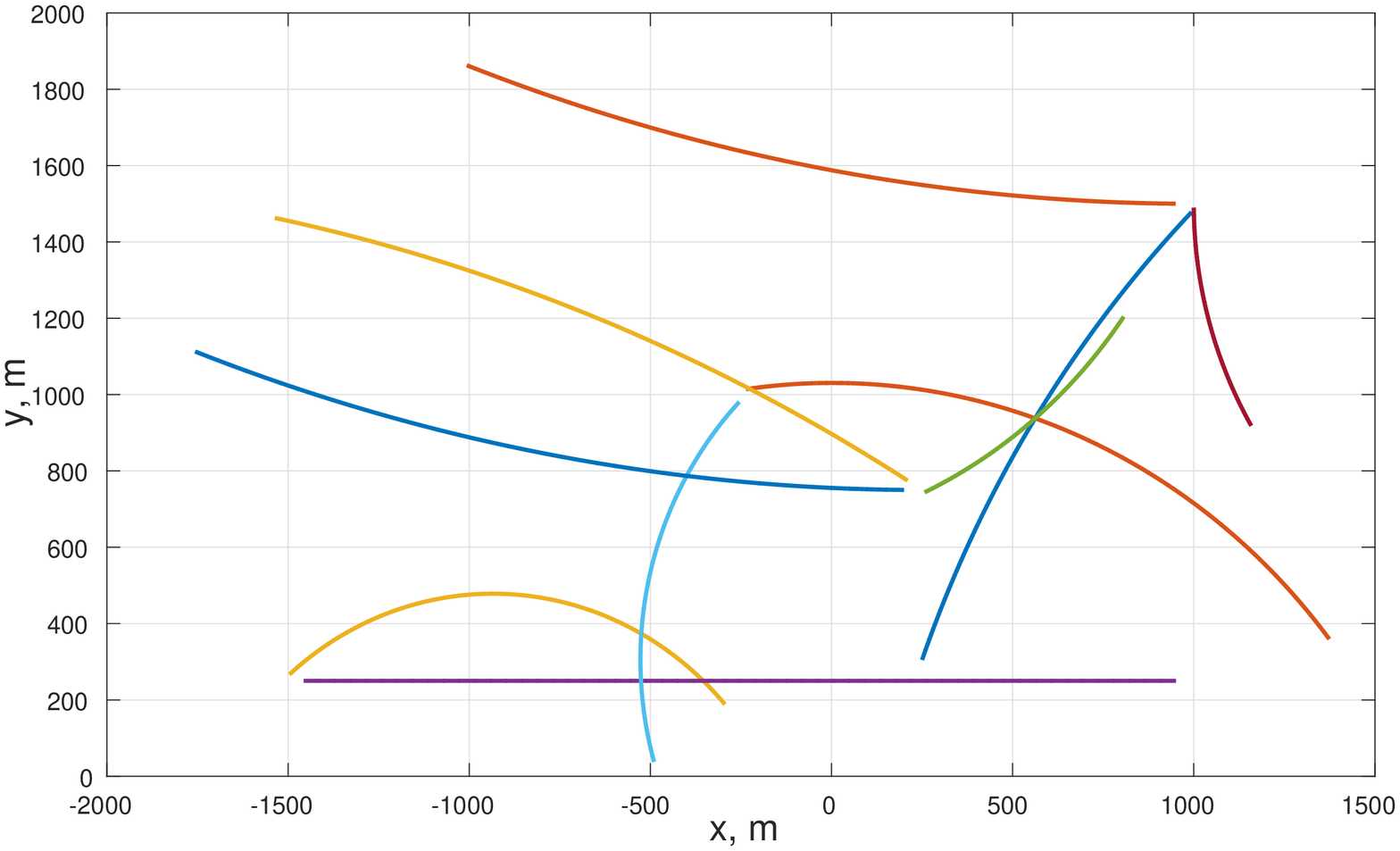}}
\end{center}
\vspace*{-.2in}
\caption{Cartesian coordinates for true target positions.}
\label{ex1_act_traj}
\end{figure}

\begin{figure}[htb]
\begin{center}
\hspace*{-.3in}  \resizebox{3.55in}{!}{\includegraphics{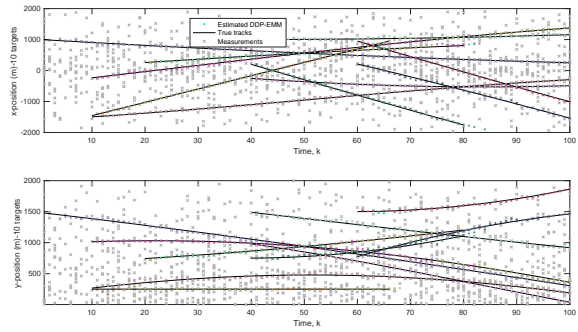}}
 \end{center}
 \vspace*{-.2in}
\caption{Actual and estimated  (a) $x_k$ and (b) $y_k$ coordinates 
using DDP-EMM (all colors)}
\label{ex1_est}
 \end{figure}

\begin{figure}[htb]
\begin{center}
\resizebox{3.5in}{2in}{\includegraphics{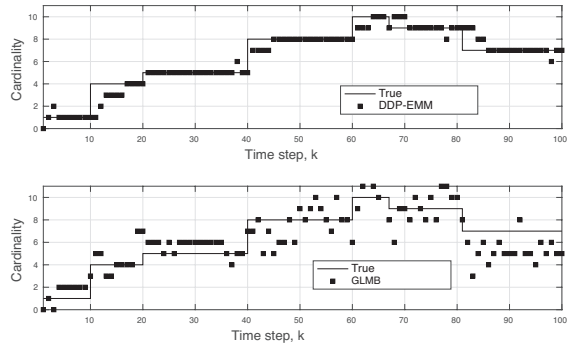}}
\end{center}
\vspace*{-.2in}
\caption{Estimated cardinality  using DDP-EMM (top) and GLMB.\vspace*{-.2in}}
\label{ex1_card}
\end{figure}

\begin{figure}[htb]
\begin{center}
\resizebox{3.5in}{2in}{\includegraphics{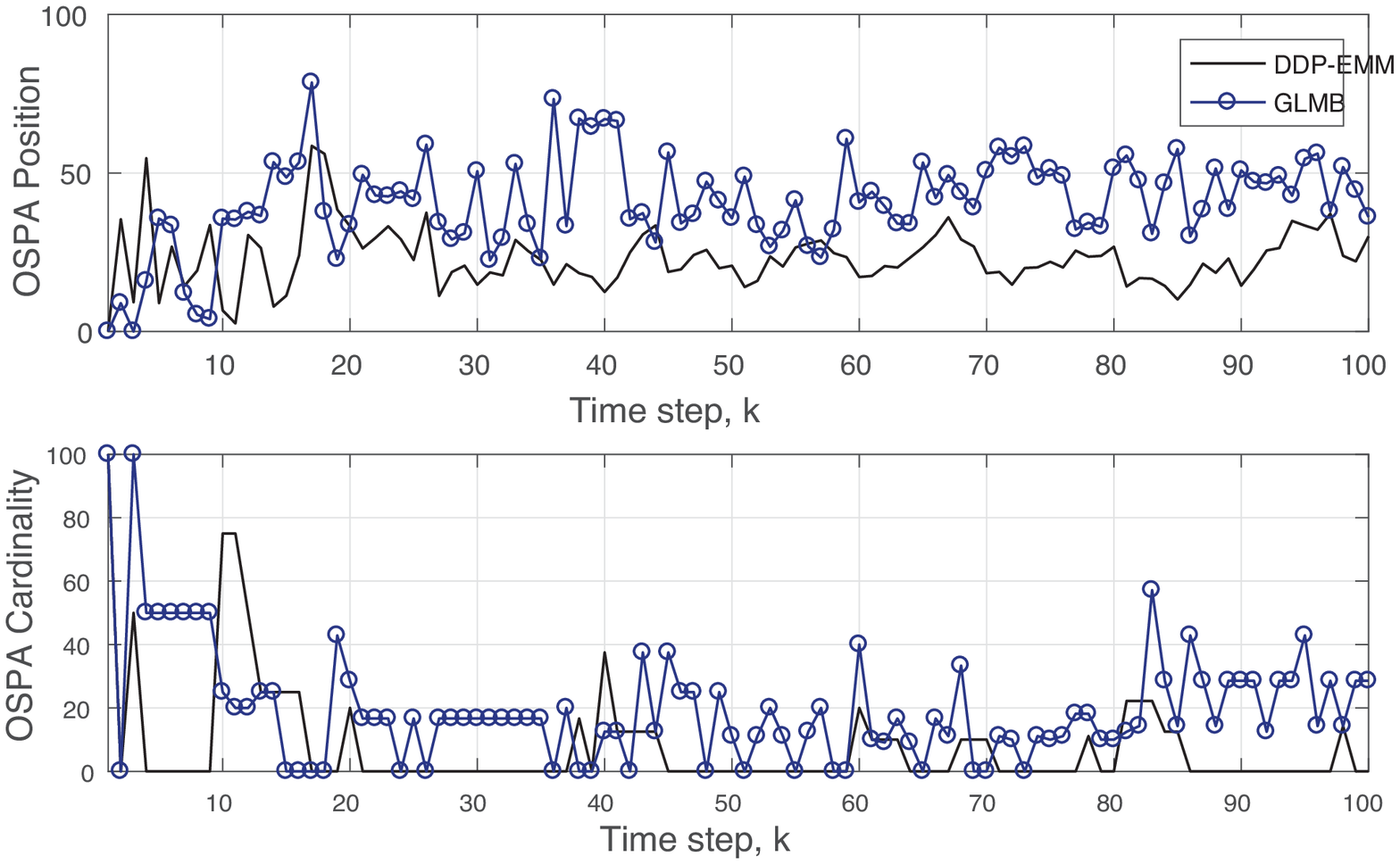}}
\end{center}
\vspace*{-.2in}
\caption{OSPA position and cardinality:  DDP-EMM, GLMB.\vspace*{-.2in}}
\label{ex1_ospa}
\end{figure}

\paragraph{DDP-EMM and GLMB Comparison} 
We compare the DDP-EEM  with the GLMB filter using the same trajectories 
used in  \cite{vo2017} (see Section IV.B.~on the non-linear 
numerical studies example) for a maximum of 10 moving targets.
The times each target enters and leaves
the scene are listed in Table \ref{existence} and 
Fig.~\ref{ex1_act_traj} depicts the true 2-D position of the targets.
The estimated $x_k$ and $y_k$ coordinates, obtained using
the DDP-EMM and GLMB methods, are compared in  Figure  \ref{ex1_est}.
The estimated cardinality 
and OSPA metric for position and cardinality 
are shown in Fig.~\ref{ex1_card} and \ref{ex1_ospa}, respectively.
 All comparisons demonstrate the increased performance
offered by the new DDP-EMM approach; part of this
may be attributed to the  approximations used by the GLMB  to update the tracks.
 In Fig.~\ref{ex1_card},  the GLMB is shown to  overestimate
  the target cardinality when compared to the DDP-EMM, 
 showing the elimination of the posterior cardinality bias. \\
 
\begin{table}[htb]
\begin{center}
\begin{tabular}{|l|l|l|l|}  \hline
Object& Presence   & Object & Presence  \\ \hline
 Object 1&$0\leq k\leq100$ & Object 6 & $40\leq k\leq100$   \\
 Object 2 & $10\leq k\leq100$  & Object 7  & $40\leq k\leq100$   \\
Object 3 & $10\leq k\leq100$  & Object 8 & $40\leq k\leq80$  \\
 Object 4 & $10\leq k\leq60$  & Object 9& $60\leq k\leq100$  \\
Object  5 & $20\leq k\leq80$  & Object 10 & $60\leq k\leq100$  \\ \hline
\end{tabular}
\end{center}
\vspace*{-.1in}
\caption{Time step of targets entering and leaving the scene.\vspace*{-.2in}}
\label{existence}
\end{table}

\begin{figure}[t]
\begin{center}
\resizebox{2.7in}{1.8in}{\includegraphics{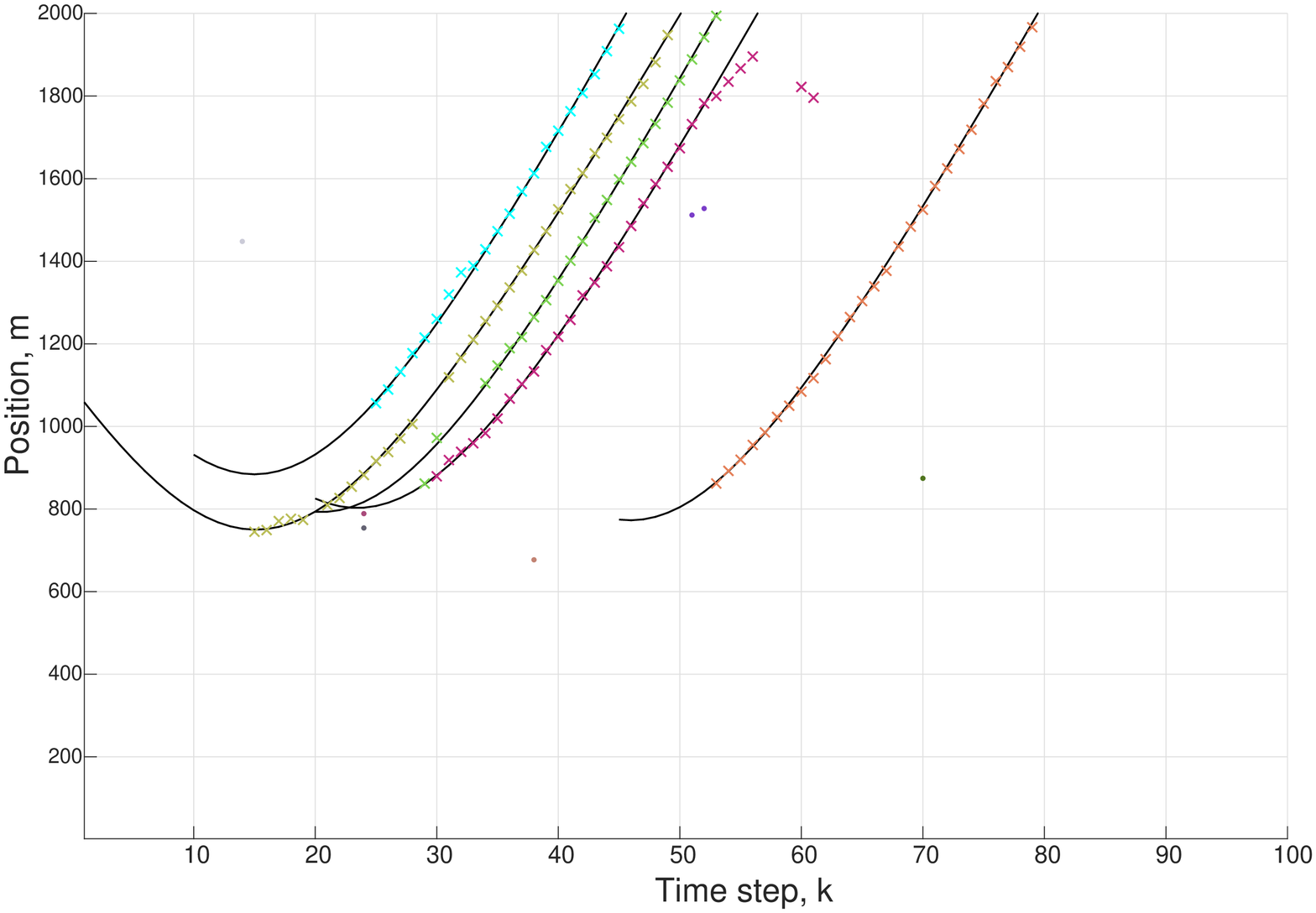}}
\end{center}
\vspace*{-.2in}
\caption{Estimated position using DDP-EM.\vspace*{-.2in}}
\label{ex2_loc}
\end{figure}

\begin{figure}[htb]
\begin{center}
\resizebox{3.8in}{2in}{\includegraphics{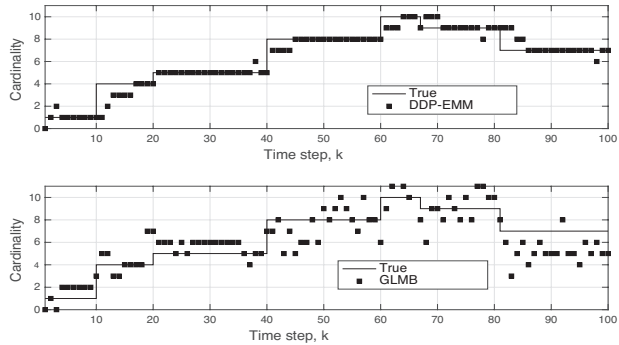}}
\end{center}
\vspace*{-.2in}
\caption{Estimated cardinality: DDP-EMM (top), GLMB.\vspace*{-.2in}}
\label{ex2_card}
\end{figure}

\begin{figure}[htb]
\begin{center}
\hspace*{-8mm} \resizebox{3.8in}{2in}{\includegraphics{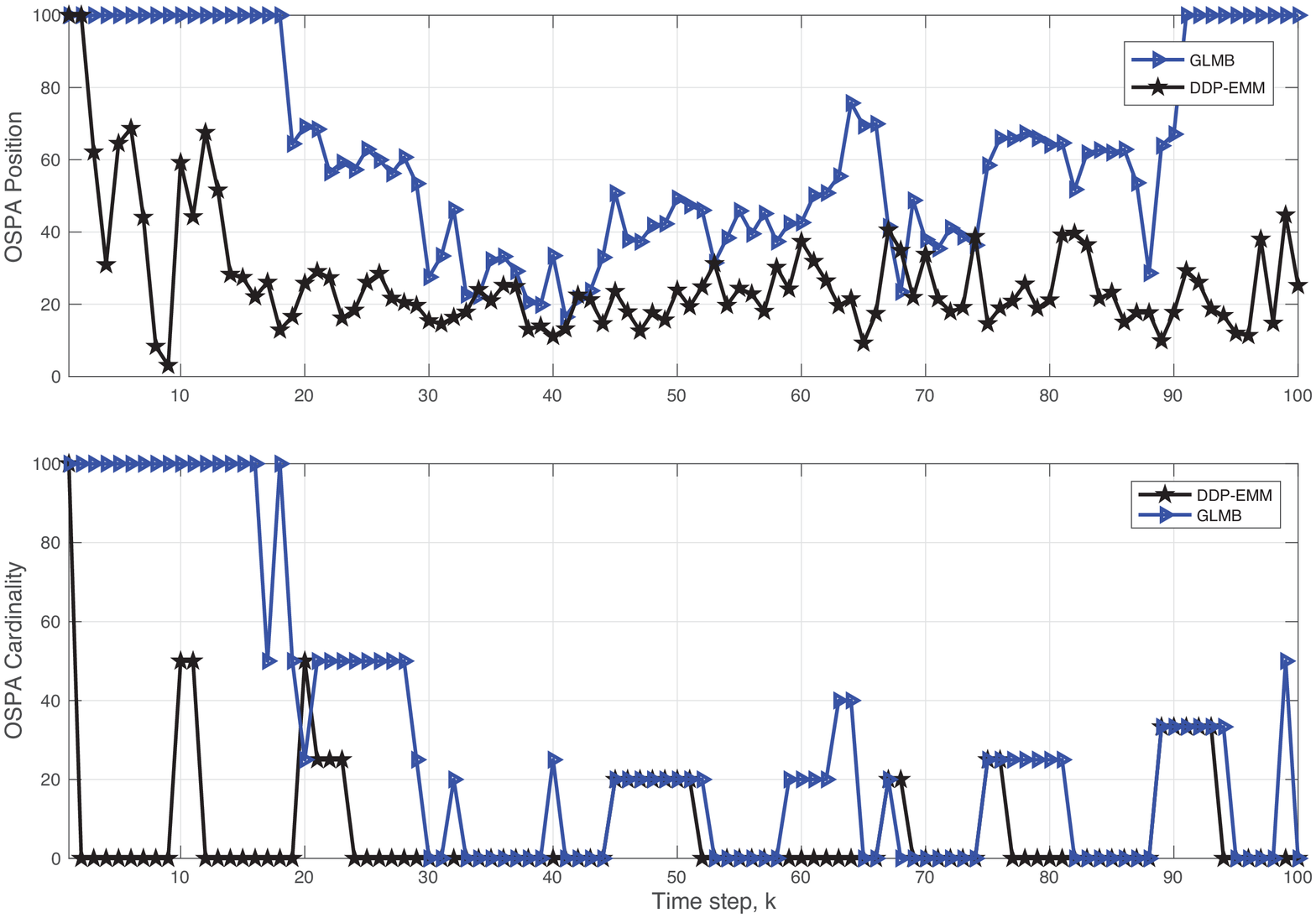}}
\end{center}
\vspace*{-.2in}
\caption{OSPA position  and cardinality: DDP-EMM, GLMB.}
\label{ex2_ospa}
\end{figure}

\paragraph{Close Proximity}
We consider a more complex scenario, where the targets
are moving in close  proximity to  each other.
A maximum of 5 targets  enter the scene
at different time steps but follow the same trajectory.
As a result, they are expected to have the same  position but at different time steps.  
The  5 targets enter the scene at corresponding 
time steps $k\eqq0$,  $k\eqq5$, $k\eqq20$, $k\eqq30$, and $k\eqq40$;
they leave the scene at corresponding  time steps
$k\eqq70$,  $k\eqq100$, $k\eqq100$, $k\eqq45$, and $k\eqq80$.
For this example, the NIW distribution  used  $\nu\eqq100$ and  concentration 
 parameter prior $\Gamma(\alpha; 1, 0.3)$.
 The estimated target position using DDP-EEM is  shown in Fig.~\ref{ex2_loc}.
 The comparison between the DDP-EEM and GLMB
 for the estimated cardinality and OSPA 
 metric are shown in Fig.~\ref{ex2_card}  and \ref{ex2_ospa}, respectively.
 As demonstrated, the DDP-EMM   performs much higher than the GLMB
  for closely-spaced targets. 


\begin{figure}[ht]
\centering
\hspace*{-4mm} \resizebox{3.8in}{2in}{\includegraphics{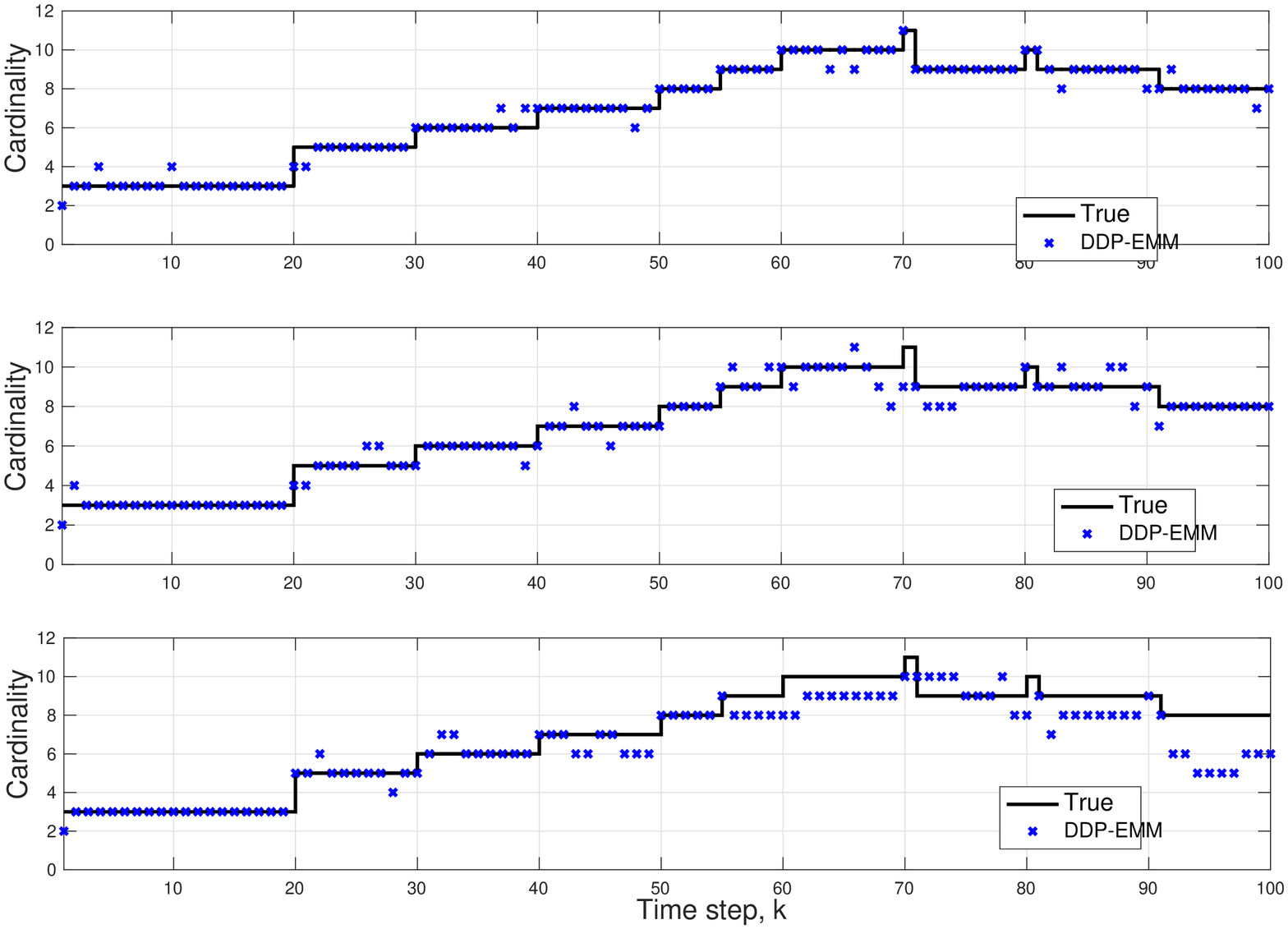}}
\vspace*{-.2in}
\caption{Cardinality estimation using DDP-EMM for -3 dB (top),
-5 dB (middle), and -10 dB SNR (bottom).\vspace*{-.2in}}
 \label{c-snr}
\end{figure}

\begin{figure}[h]
\centering
\hspace*{-4mm}  \resizebox{3.8in}{2in}{\includegraphics{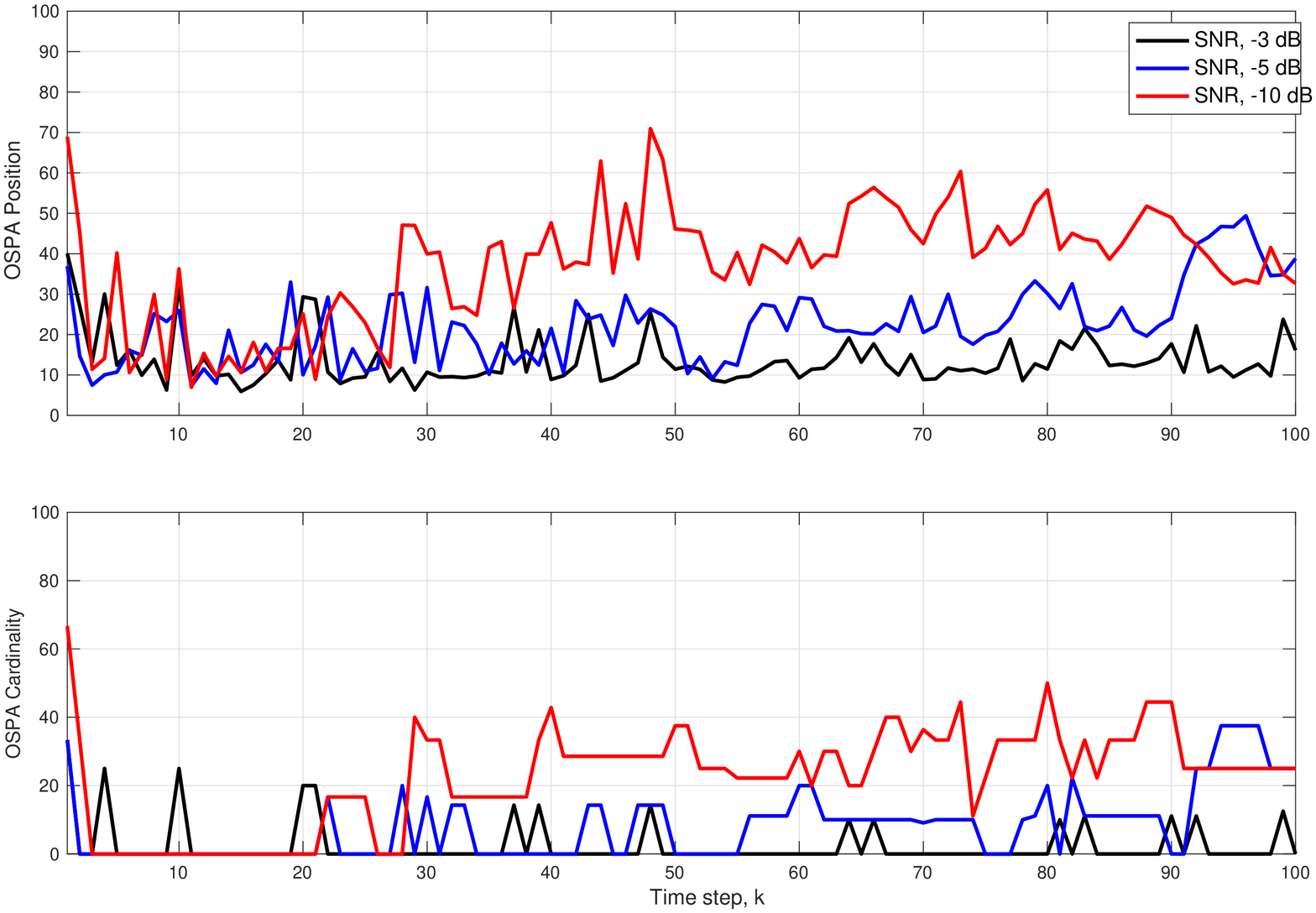}}
\vspace*{-.1in}
\caption{OSPA position (top) and cardinality (bottom)  using 
DDP-EMM  for  -3 dB, -5 dB and -10 dB  SNR.}
\label{o-snr}
\end{figure}

\paragraph{Varying SNR}
In this example, we assume that there are 11 targets with  
$\omega_k\eqq 0$ and that only range  measurements are available.
 We demonstrate the performance of the DDP-EMM approach for 
 SNR values -3, -5 and -10 dB.  For the simulations, 
  $\mu_0\eqq 0$ and $\nu\eqq100$ were used
 for the NIW distribution  and the concentration prior  was $\Gamma(\alpha; 1, 0.2)$.
  The cardinality performance for decreasing SNR is demonstrated
  in Fig.~\ref{c-snr}.   As expected, the performance of the DDP-EMM 
  decreases as the SNR decreases; however, the  
  correct cardinality of the states was obtained most of times. 
Fig.~\ref{o-snr} depicts the decrease in 
 DDP-EMM OSPA performance as the SNR decreases.
 %


\begin{figure}[tbh]
\begin{center}
    \resizebox{3.8in}{2in}{\includegraphics{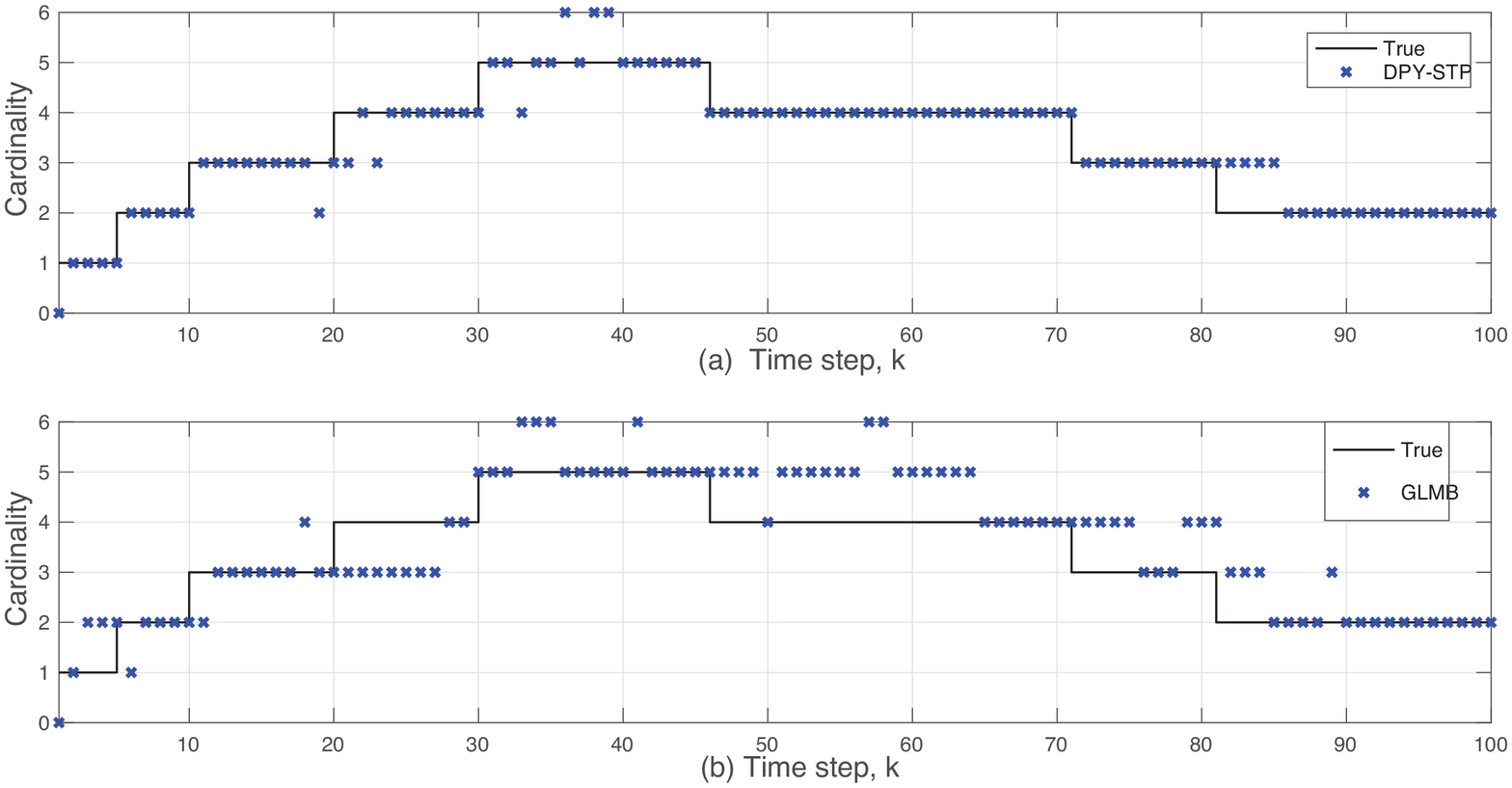}}
  \end{center}
   \vspace*{-.2in}
 \caption{Cardinality estimation: (a)  DPY-STP (b) GLMB}.
\label{card_DPY}
\end{figure}

\begin{figure}[tbh]
\begin{center}
  \resizebox{3.8in}{2in}{\includegraphics{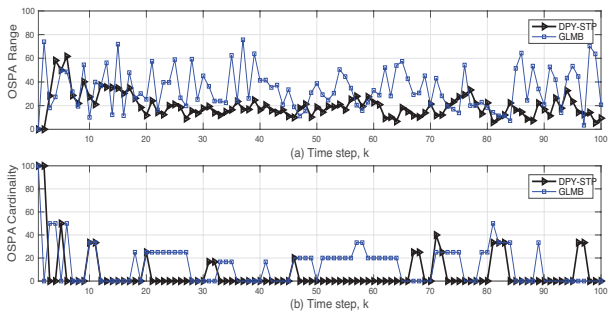}}
  \end{center}
  \vspace*{-.3in}
  \caption{OSPA range \& cardinality: DPY-STP, GLMB.\vspace*{-.1in}}
 \label{fig:ospa}
\end{figure}

\paragraph{DPY-STP and GLMB Comparison}
The higher performance of the DPY-STP is demonstrated 
and compared to the GLMB using a maximum of 5 targets. 
This is shown in Fig.~\ref{card_DPY} and 
Fig.~\ref{fig:ospa} with the estimated cardinality and 
the OSPA range and cardinality, respectively, for the two methods.

\begin{figure}[t]
\centering
\subfloat[ ]{ 
 \resizebox{3.4in}{2in}{\includegraphics{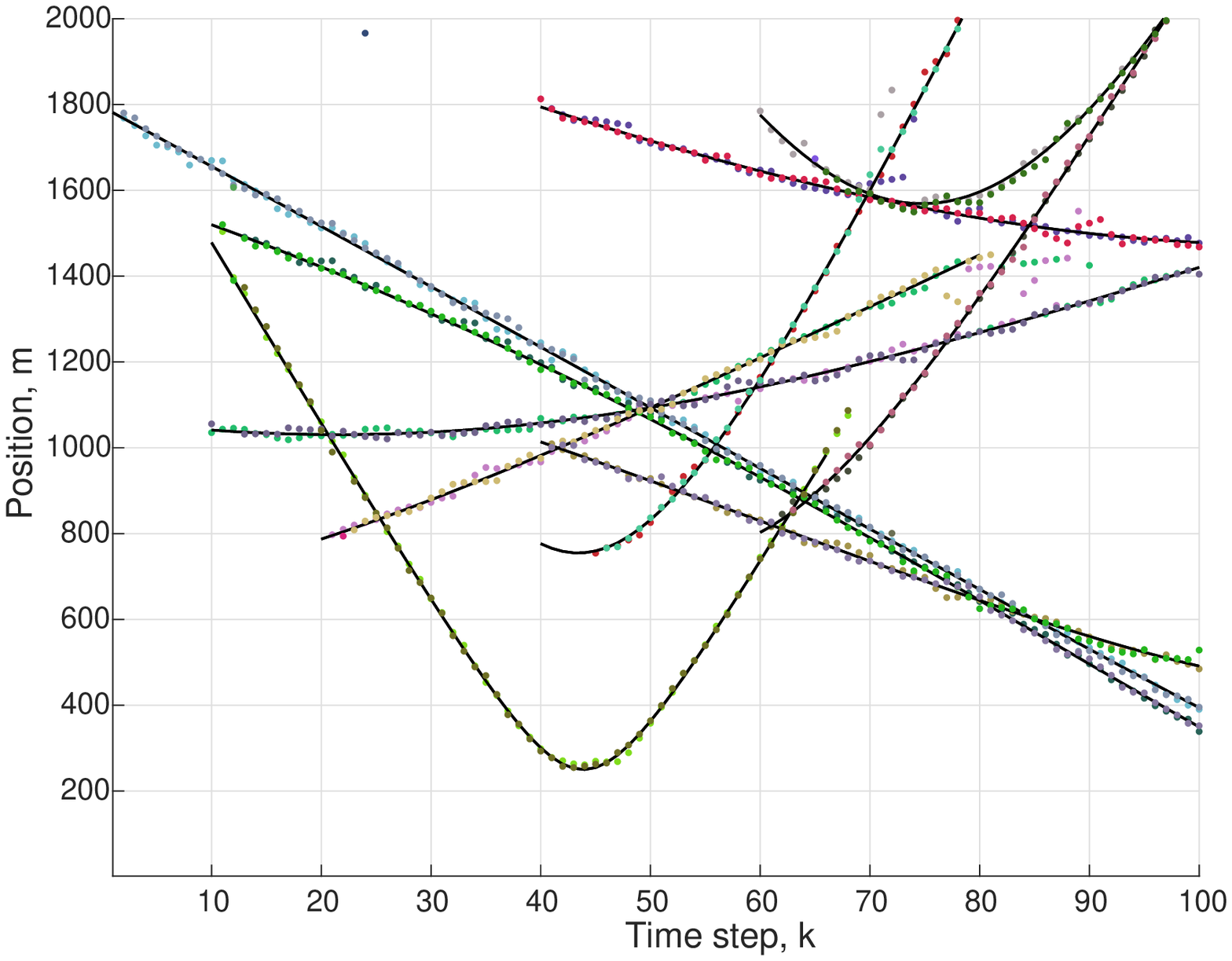}}}   \\[-1mm]
 \subfloat[ ]{ 
  \resizebox{3.4in}{2in}{\includegraphics{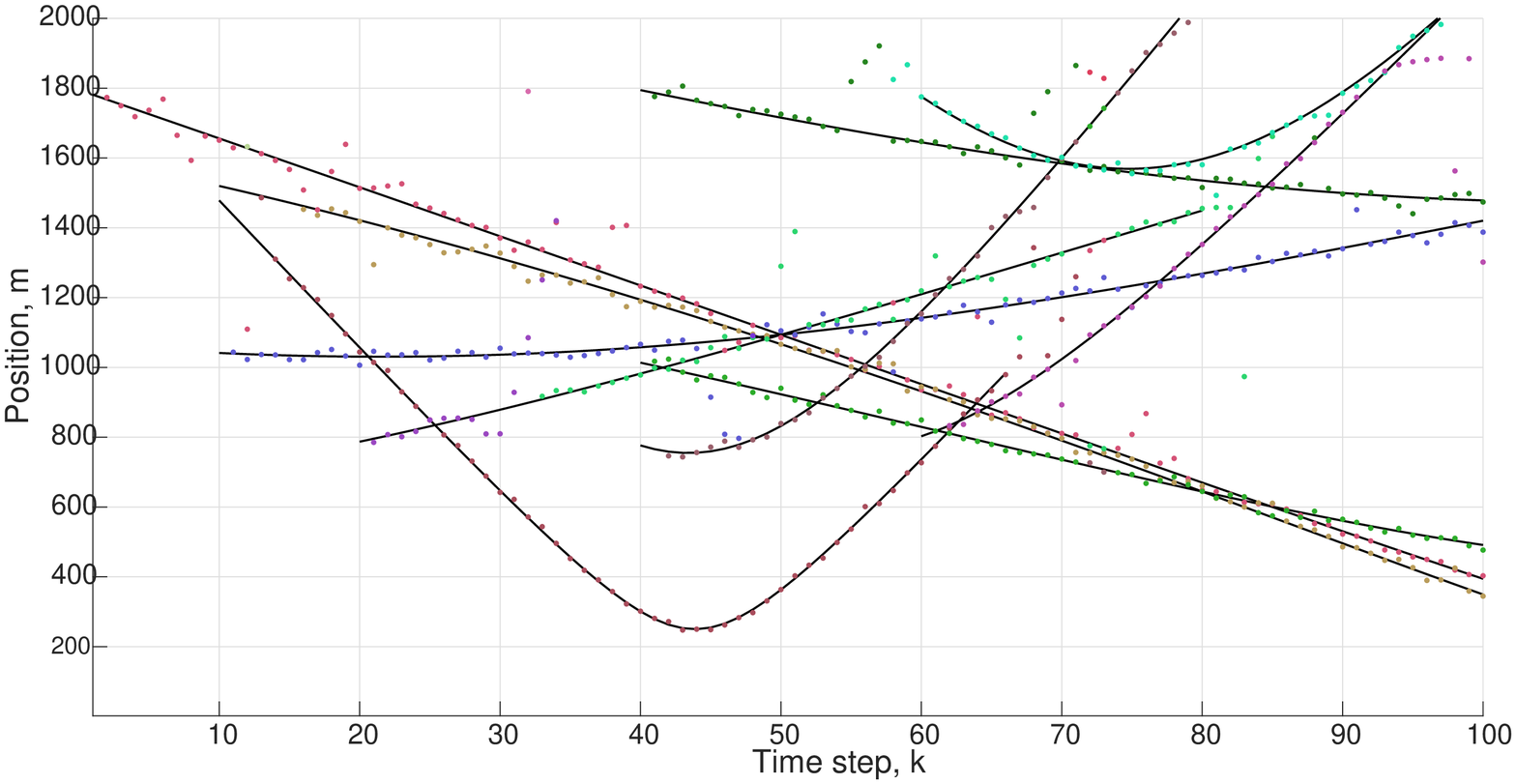}}}
  \vspace*{-,1in}
 \caption{Position estimation: (a)  DPY-STP (b) DDP-EMM.}
  \label{locpy}
 \end{figure}

\begin{figure}[tbh]
 \begin{center}
 \resizebox{3.5in}{1.8in}{\includegraphics{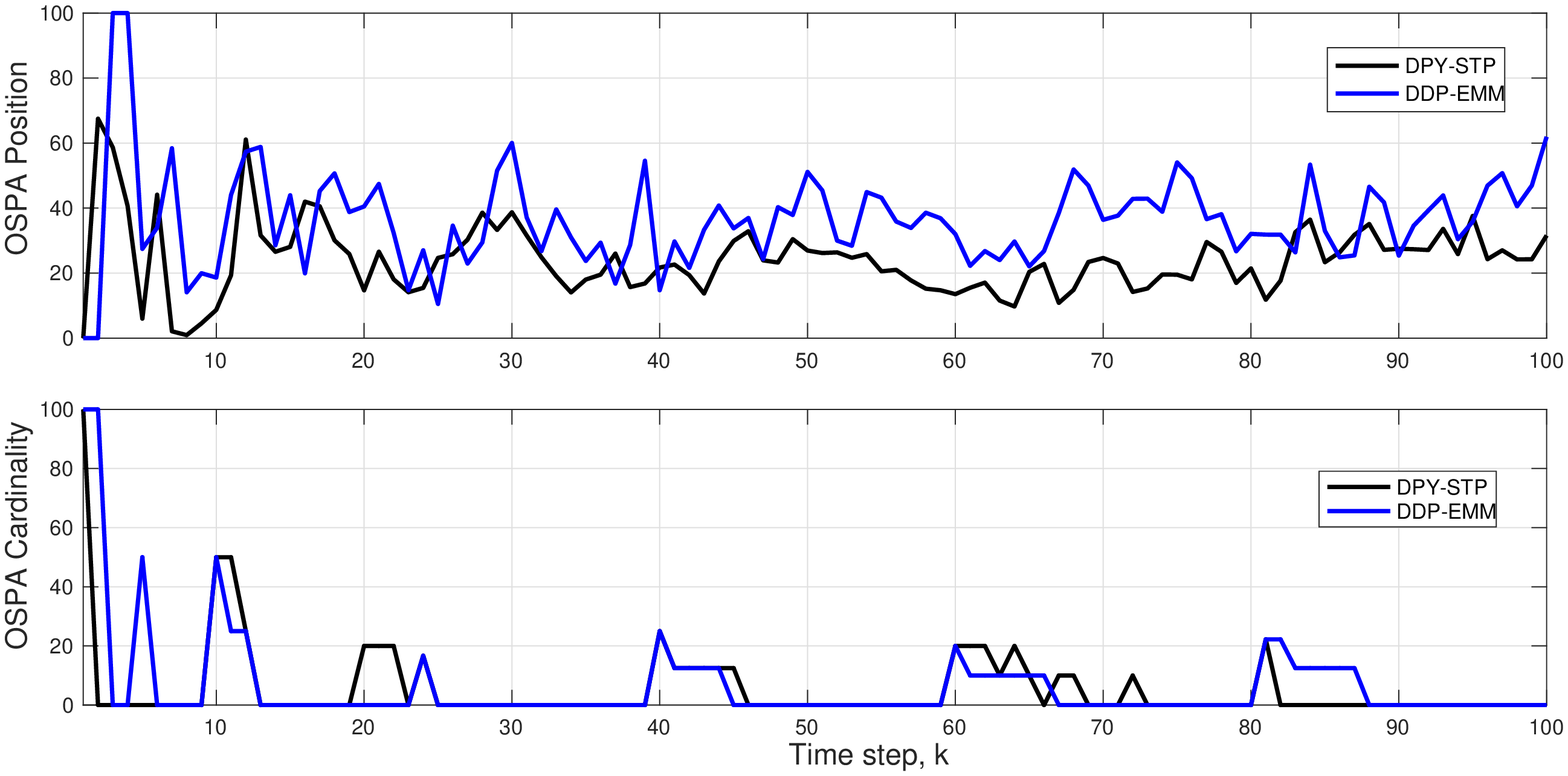}}
 \end{center}
  \vspace*{-.2in}
\caption{OSPA  position \& cardinality:  
DPY-STP, DDP-EMM.}
 \label{pydp}
\end{figure}

\paragraph{DPY-STP and DDP-EMM Comparison}
As discussed in Section \ref{all_DPYP}, the  DPY is  
better matched to  the tracking a  TV number of objects 
than the DDP.  This is demonstrated by comparing
the DPY-STP and DDP-EMM methods 
using a maximum number of 10 targets.
The simulations used  $\mu_0\eqq 0$ and $\nu\eqq100$ 
 for the NIW distribution; parameters 
 $\alpha$ and $d$ were selected using 
 $\Gamma(\alpha; 1, 0.1)$.
 The estimated position using the two methods 
 is shown in Fig.~\ref{locpy}.
 The increased performance of the DPY-STP is 
 shown using the   OSPA comparison for position and cardinality 
  in Fig.~\ref{pydp}.

\section{Conclusion} 
We presented novel families of Bayesian 
nonparametric processes to capture
the computational and inferential needs when tracking 
a TV number of  objects.   The  proposed models offer improvements
in tracking performance, time efficiency, and implementation cost.  
They exploit the dependent Dirichlet and dependent Pitman-Yor processes to
model object priors to efficiently track object labels,  cardinality and trajectories.
Furthermore, MCMC implementation of the proposed
tracking algorithms successfully verifies the
simplicity and accuracy of proposed methods.


\acks{The authors would like to thank the anonymous referees, an Associate
Editor and the Editor for their constructive comments that improved the
quality of this paper.This work was supported in part by Grant AFOSR FA9550-17-1-0100 }



\appendix
\section*{Appendix A.}
\label{app:theorem}



In this appendix we prove the equation 14:
Section~4.2:

\noindent

{\bf Proof}. Proof of \ref{Gibbs} follows the standard Bayesian nonparametric methods. We know that the base measure in DP($\alpha$, H) is the mean of the Dirichlet prior. The following lemma generalizes this fact.
\begin{lemma}(Ferguson 1973, \cite{ferg1973})
\label{mean}
If $G \sim \text{DP}(\alpha, H)$ and $f$ is any measurable function, then 
\begin{equation}
\mathbb{E}\Big[ \int f(\theta) dG(\theta)\Big] = \int f(\theta) dH(\theta) \notag
\end{equation}
\end{lemma}
Suppose that $A$ and $B$ are measurable sets. 
\begin{flalign}
P(\theta_{\ell,k}\in A , \bz_{\ell, k} \in B| \theta_{-\ell,k}, \bz_{-\ell, k}) = &\mathbb{E} \big[ \mathbbm{1}_{\theta_{\ell,k}}(A)\mathbbm{1}_{\bz_{\ell, k}}(B)| \theta_{-\ell,k}, \bz_{-\ell, k}\big]\label{d1}\\
=&\mathbb{E}\Big[\mathbb{E} \big[ \mathbbm{1}_{\theta_{\ell,k}}(A)\mathbbm{1}_{\bz_{\ell, k}}(B)| G, \theta_{-\ell,k}, \bz_{-\ell, k}\big] | \theta_{-\ell,k}, \bz_{-\ell, k}\Big]\label{d2}\\
= &\mathbb{E}\Big[ \int \mathbbm{1}_{\theta_{\ell,k}}(A)\mathbbm{1}_{\bz_{\ell, k}}(B) p(\bz_{\ell,k}|\theta_{\ell,k}, \bx_{\ell,k})d\bz_{\ell, k} dG({\theta_{\ell,k}}|{\theta_{-\ell,k}})\Big]\label{d3}
\end{flalign}
where \ref{d1} follows the definition of expected value, \ref{d2} is due to the law of iterated expectations, and $G(\theta)$ in \ref{d3} is the posterior dependent Dirichlet process given in \ref{posttheta}. Using lemma \ref{mean} 
\begin{flalign}
&\mathbb{E}\Big[ \int \mathbbm{1}_{\theta_{\ell,k}}(A)\mathbbm{1}_{\bz_{\ell, k}}(B) p(\bz_{\ell,k}|\theta_{\ell,k}, \bx_{\ell,k})d\bz_{\ell, k} dG({\theta_{\ell,k}}|{\theta_{-\ell,k}})\Big] =\\
 & \int \mathbbm{1}_{\theta_{\ell,k}}(A)\mathbbm{1}_{\bz_{\ell, k}}(B) p(\bz_{\ell,k}|\theta_{\ell,k}, \bx_{\ell,k})d\bz_{\ell, k} \\ &d\Big(\sum\limits_{\Theta_k - \{\theta_{\ell,k}\}} \Pi_1 \delta_{\theta}(\theta_{\ell,k}) + \sum\limits_{\substack{\theta \in \Theta^\star_{k|k-1}\setminus \Theta\\ \theta\neq \theta_{\ell,k}}}\Pi_2 \nu({\bf{\theta}^\star_{\ell, k-1}},{\bf{\theta_{\ell, k}}}) \delta_{\theta}({\bf{\theta_{\ell,k}}}) + \Pi_3 H(\theta_{\ell,k})\Big)\notag.
 \end{flalign}
 Using the Bayes rule we have:
 \begin{flalign}
 P(\theta_{\ell,k}\in A | \theta_{-\ell,k}, \mathcal{Z}_k) = \frac{ \int_{B} P(\theta_{\ell,k}\in A , \bz_{\ell, k} | \theta_{-\ell,k}, \bz_{-\ell, k})d\bz_{\ell, k}}{\int_{\Omega}P(\theta_{\ell,k}\in A , \bz_{\ell, k} | \theta_{-\ell,k}, \bz_{-\ell, k})d\bz_{\ell, k}}
 \end{flalign}
 and this concludes the claim in \ref{Gibbs}.

\vskip 0.2in
\bibliography{main}

\end{document}